\batchmode
\makeatletter
\def\input@path{{\string"E:/Trabajo Angel/Mis articulos/PMBM_Targets/Accepted/\string"}}
\makeatother
\documentclass[british,english]{IEEEtran}
\usepackage[T1]{fontenc}
\usepackage[latin9]{inputenc}
\usepackage{float}
\usepackage{amsmath}
\usepackage{amsthm}
\usepackage{amssymb}
\usepackage{graphicx}
\usepackage{nomencl}

\providecommand{\makenomenclature}{\makeglossary}
\makenomenclature

\makeatletter

\providecommand{\tabularnewline}{\\}
\floatstyle{ruled}
\newfloat{algorithm}{tbp}{loa}
\providecommand{\algorithmname}{Algorithm}
\floatname{algorithm}{\protect\algorithmname}

\theoremstyle{plain}
\newtheorem{thm}{\protect\theoremname}
\theoremstyle{definition}
\newtheorem{example}[thm]{\protect\examplename}
\theoremstyle{plain}
\newtheorem{cor}[thm]{\protect\corollaryname}
\theoremstyle{plain}
\newtheorem{prop}[thm]{\protect\propositionname}

\usepackage{cite} 
\allowdisplaybreaks 
\usepackage[margin=8pt,font=footnotesize]{caption}
\usepackage{algorithm}
\usepackage{algpseudocode}

\makeatother

\usepackage{babel}
\addto\captionsbritish{\renewcommand{\algorithmname}{Algorithm}}
\addto\captionsbritish{\renewcommand{\corollaryname}{Corollary}}
\addto\captionsbritish{\renewcommand{\examplename}{Example}}
\addto\captionsbritish{\renewcommand{\propositionname}{Proposition}}
\addto\captionsbritish{\renewcommand{\theoremname}{Theorem}}
\addto\captionsenglish{\renewcommand{\corollaryname}{Corollary}}
\addto\captionsenglish{\renewcommand{\examplename}{Example}}
\addto\captionsenglish{\renewcommand{\propositionname}{Proposition}}
\addto\captionsenglish{\renewcommand{\theoremname}{Theorem}}
\providecommand{\corollaryname}{Corollary}
\providecommand{\examplename}{Example}
\providecommand{\propositionname}{Proposition}
\providecommand{\theoremname}{Theorem}

\begin{document}

\title{Poisson multi-Bernoulli mixture filter: direct derivation and implementation}

\author{Ángel F. García-Fernández, Jason L. Williams, Karl Granström, Lennart
Svensson \thanks{© 2018 IEEE. Personal use of this material is permitted. Permission from IEEE must be obtained for all other uses, in any current or future media, including reprinting/republishing this material for advertising or promotional purposes, creating new collective works, for resale or redistribution to servers or lists, or reuse of any copyrighted component of this work in other works.

Ángel F. García-Fernández is with the Department of Electrical Engineering and Electronics, University of Liverpool, Liverpool L69 3GJ, United Kingdom (email: angel.garcia-fernandez@liverpool.ac.uk). Jason L. Williams is with the National Security and ISR Division, Defence Science and Technology Group, Edinburgh SA 5111, Australia and also with the School of Electrical Engineering and Computer Science, Queensland University of Technology, Brisbane, QLD 4000, Australia (email: Jason.Williams@dst.defence.gov.au). Karl Granström and Lennart Svensson are with the Department of Electrical Engineering, Chalmers University of Technology, SE-412 96 Gothenburg, Sweden (emails: \{karl.granstrom, lennart.svensson\}@chalmers.se).}}
\maketitle
\begin{abstract}
We provide a derivation of the Poisson multi-Bernoulli mixture (PMBM)
filter for multi-target tracking with the standard point target measurements
without using probability generating functionals or functional derivatives.
We also establish the connection with the $\delta$-generalised labelled
multi-Bernoulli ($\delta$-GLMB) filter, showing that a $\delta$-GLMB
density represents a multi-Bernoulli mixture with labelled targets
so it can be seen as a special case of PMBM. In addition, we propose
an implementation for linear/Gaussian dynamic and measurement models
and how to efficiently obtain typical estimators in the literature
from the PMBM. The PMBM filter is shown to outperform other filters
in the literature in a challenging scenario. 
\end{abstract}


\section{Introduction}

Multiple target tracking (MTT) is an important problem with many
different uses, for example, in aerospace applications, surveillance,
air traffic control, computer vision and autonomous driving \cite{Blackman04,Koch11,Garcia09,Bhaskar15,Petrovskaya09,Kunz15}.
In MTT, a variable and unknown number of targets appear, move and
disappear from a scene of interest. At each time step, these targets
are observed through noisy measurements, possibly coming from multiple
sensors \cite{Fantacci16,Meyer17}, and the aim is to infer where
the targets are at each time step. 

The random finite set (RFS) framework is widely used to model this
problem in a Bayesian way \cite{Mahler_book14}. Here, the usual set-up
is to consider the state of the system at the current time as a set
of targets. There are a variety of dynamic models \cite{Li03} for
this set of targets but it is usually assumed that it evolves in time
according to a Markov process, which also accounts for target births/deaths.
There are also different widely used measurement models, for example,
standard (point target) \cite{Mahler_book14}, extended target \cite{Swain10,Granstrom12}
or track-before-detect \cite{Angel13,Davey08} measurement models.

As in any Bayesian setting, the information of interest about the
targets at the current time step is contained in the (multitarget)
density of the current set of targets given present and past measurements.
In theory, this density can be computed via the prediction and update
steps of the Bayesian filtering recursion. However, in general, this
computation is intractable and general, computationally expensive
approximations such as particle filters should be used \cite{Ristic_book04}.
Nevertheless, as we explain next, there are families of multi-target
densities that are conjugate prior for some models that enable easier
and more efficient computation. 

In Bayesian probability theory, a family of probability distributions
is conjugate for a given likelihood function if the posterior distribution
for any member of this family also belongs to the same family \cite{Robert_book07}.
In MTT filtering, it is especially useful for computational reasons
to consider conjugate priors in which the posterior distributions
can be written explicitly in terms of single target Bayesian updates,
which might not admit a closed-form expression \cite{Vo13,Williams15b}.
Additionally, in MTT, it is convenient to introduce conjugacy for
the prediction step. That is, a multi-target density is conjugate
with respect to a dynamic model if the same family is preserved after
performing the prediction step. This conjugacy property for the prediction
and update steps is quite important in the RFS context as it allows
the posterior to be written in terms of single target predictions
and updates, which are much easier to compute/approximate than full
multi-target predictions and updates. Due to this important characteristic,
in general, when we refer to MTT conjugacy, we are referring to a
family of distributions which is closed under both prediction and
update steps. Note that, in MTT, we are generally dealing with conjugate
prior mixtures in which the number of mixture components can grow,
due to the data association. This implies that the conjugate prior
does not have a fixed dimensional sufficient statistic, even if the
single target densities have it.

We proceed to describe the two conjugate priors in the literature
for the standard (point target) measurement model, in which the set
of measurements at a given time comprises clutter and one or zero
measurements per target. The first conjugate prior consists of the
union of a Poisson process and a multi-Bernoulli mixture (PMBM) \cite{Williams15b}.
Importantly, the multi-Bernoulli mixture, which considers all the
data association hypotheses, can be implemented efficiently using
a track-oriented multiple hypotheses tracking (MHT) formulation \cite{Kurien_inbook90}.
The Poisson part considers all targets that have never been detected
and enables an efficient management of the number of hypotheses covering
potential targets \cite{Williams15b}. The second conjugate prior
was presented for labelled targets in \cite{Vo13}. In the usual radar
tracking case, in which targets do not have a unique ID, labels are
artificial variables that are added to the target states with the
objective of estimating target trajectories \cite{Angel09,Angel13,Vo13,Vo14,Aoki16}.
With them, we can also obtain conjugate priors, as in the $\delta$-generalised
labelled multi-Bernoulli ($\delta$-GLMB) filter \cite{Vo13,Vo14}.

The PMBM filter in \cite{Williams15b}, which is based on the previously
mentioned conjugate prior, was derived by using probability generating
functionals (PGFLs) and functional derivatives \cite{Mahler07}. These
are very important tools for deriving RFS filters, such as the probability
hypothesis density (PHD) or cardinalised PHD (CPHD) filters \cite{Mahler03,Mahler07}.
However, non-PGFL derivations are also useful as they can provide
insights about the structure of the filter and make the understanding
of the filter accessible to more researchers, as was done in \cite{Angel15_d}
for the PHD and CPHD filters. 

The main aim of this paper is to make the PMBM filter accessible to
a wider audience from a theoretical and practical point of view. In
order to do so, we make the following contributions: 1. In Section
\ref{sec:Conjugate-priors-proof}, we provide a derivation of the
PMBM filter for point measurements that does not rely on PGFLs or
functional derivatives, improving the accessibility of these results
and providing more insight into the structure of the solution. 2.
In Section \ref{sec:Connection-delta-GLMB}, we show that the $\delta$-GLMB
(multi-target) density can be seen as a special case of a PMBM on
a labelled state space, and discuss the benefits of the PMBM form.
3. Section \ref{sec:Linear-Gaussian-implementation} proposes an implementation
of the PMBM filter for linear/Gaussian dynamic and measurement models.
4. In Section \ref{sec:Estimation}, we provide tractable methods
for obtaining the estimators used in MHT and the $\delta$-GLMB filter
using the PMBM distribution form. We also provide a third estimator
that improves performance for high probability of detection. 5. Finally,
Section \ref{sec:Simulations} demonstrates the PMBM implementation
on a challenging scenario, comparing performance between the three
estimators and other multi-target filters.

\section{Bayesian filtering with random finite sets\label{sec:Bayesian-filtering-with}}

In Section \ref{subsec:Filtering-recursion}, we review the Bayesian
filtering recursion with random finite sets. In Section \ref{subsec:Standard-point-target-measurement},
we present the likelihood function for the standard point target measurement
model.

\subsection{Filtering recursion\label{subsec:Filtering-recursion}}

In this section we review the Bayesian filtering recursion with RFSs,
which consists of the usual prediction and update steps. As we only
need to consider one prediction and update step, we omit the time
index of the filtering recursion for notational simplicity. 

In the standard RFS framework for target tracking, we have a single
target state $x\in\mathbb{R}^{n_{x}}$ and a multi-target state $X\in\mathcal{F}\left(\mathbb{R}^{n_{x}}\right)$,
where $X$ is a set whose elements are single target state vectors
and $\mathcal{F}\left(\mathbb{R}^{n_{x}}\right)$ denotes the space
of all finite subsets of $\mathbb{R}^{n_{x}}$. In the update step,
the state is observed by measurements that are represented as a set
$Z\in\mathcal{F}\left(\mathbb{R}^{n_{z}}\right)$. Given a prior (multi-target)
density $f\left(\cdot\right)$ and the (multi-target) density $l(Z|X)$
of the measurement $Z$ given the state $X$, the posterior multi-target
density of $X$ after observing $Z$ is given by Bayes' rule \cite{Mahler03}
\begin{align}
q(X) & =\frac{l(Z|X)f(X)}{\rho(Z)}\label{eq:Bayes_rule}
\end{align}
where the normalising constant is
\begin{align}
\rho(Z) & =\int l(Z|X)f(X)\delta X\label{eq:PDF_measurement}\\
 & =\sum_{n=0}^{\infty}\frac{1}{n!}\int l\left(Z\left|\left\{ x_{1},...,x_{n}\right\} \right.\right)\nonumber \\
 & \quad\times f\left(\left\{ x_{1},...,x_{n}\right\} \right)d\left(x_{1},...,x_{n}\right).\label{eq:set_integral_measurement}
\end{align}

The Bayesian filtering recursion is completed with the prediction
step. Given a posterior density $q\left(\cdot\right)$, the prior
density $\omega\left(\cdot\right)$ at the next time step is given
by the Chapman-Kolmogorov equation
\begin{align}
\omega\left(X'\right) & =\int\gamma\left(X'|X\right)q\left(X\right)\delta X\label{eq:prediction_step}
\end{align}
where $X'\in\mathcal{F}\left(\mathbb{R}^{n_{x}}\right)$ denotes the
state at the next time step and $\gamma\left(X'|X\right)$ is the
transition density of the state $X'$ given the state $X$. We consider
the conventional dynamic assumptions for MTT used in the RFS framework
\cite{Mahler_book07}: at each time step, a target follows a Markovian
process such that it survives with a probability $p_{s}\left(\cdot\right)$
and moves with a transition density $g\left(\cdot\left|\cdot\right.\right)$.
New born targets follow a Poisson RFS with intensity $\lambda^{b}\left(\cdot\right)$.

\subsection{Standard point target measurement model\label{subsec:Standard-point-target-measurement}}

In this section, we provide the likelihood $l(Z|X)$ for the standard
point target measurement model, which is described next. At different
parts of this paper, we will make use of different representations
of the likelihood, which require the introduction of extra notation.
To aid the reader, a summary of this notation is found in Table \ref{tab:Notations-for-different-likelihood}. 

Given the set $X=\left\{ x_{1},...,x_{n}\right\} $ of targets, the
set $Z$ of measurements is $Z=Z^{c}\uplus Z_{1}\uplus...\uplus Z_{n}$
where $Z^{c}$, $Z_{1}$,..., $Z_{n}$ are independent sets, $Z^{c}$
is the set of clutter measurements, $Z_{i}$ is the set of measurements
produced by target $i$. Symbol $\uplus$ stands for disjoint union,
which is used to represent that $Z=Z^{c}\cup Z_{1}\cup...\cup Z_{n}$
and $Z^{c},Z_{1},...,Z_{n}$ are mutually disjoint (and possibly empty)
\cite{Mahler_book14}. Set $Z^{c}$ is a Poisson point process with
intensity/PHD $c\left(\cdot\right)$. We get $Z_{i}=\emptyset$ with
probability $1-p_{d}\left(x_{i}\right)$, which corresponds to the
case where the target is not detected, and $Z_{i}=\left\{ z\right\} $
where $z$ has a density $p\left(z|x_{i}\right)$ with probability
$p_{d}\left(x_{i}\right)$, which corresponds to the case where the
target is detected. 

Using the convolution formula for multi-object densities \cite[Eq. (4.17)]{Mahler_book14},
the resulting density $l\left(\cdot|\cdot\right)$ of $Z$ given $X$
can be written as\nomenclature{$\hat{l}\left(Z|x\right)$}{Likelihood of single target $x$ for $Z$}
\begin{align}
l\left(Z|\left\{ x_{1},...,x_{n}\right\} \right) & =e^{-\lambda_{c}}\sum_{Z^{c}\uplus Z_{1}...\uplus Z_{n}=Z}\left[c\left(\cdot\right)\right]^{Z^{c}}\prod_{i=1}^{n}\hat{l}\left(Z_{i}|x_{i}\right)\label{eq:likelihood_standard}
\end{align}
\begin{align}
\hat{l}\left(Z|x\right) & =\begin{cases}
p_{d}\left(x\right)p\left(z|x\right) & Z=\left\{ z\right\} \\
1-p_{d}\left(x\right) & Z=\emptyset\\
0 & \left|Z\right|>1
\end{cases}\label{eq:l_hat}
\end{align}
where $\lambda_{c}=\int c\left(z\right)dz$ and we use the multi-object
exponential notation $\left[c\left(\cdot\right)\right]^{Z}=\prod_{z\in Z}c\left(Z\right)$,
$\left[c\left(\cdot\right)\right]^{\emptyset}=1$ \cite{Vo13}. The
notation in (\ref{eq:likelihood_standard}) means that for a given
$Z$, we perform a sum that goes through all possible sets $Z^{c}$,
$Z_{1}$,..., $Z_{n}$ that meet the requirement $Z^{c}\uplus Z_{1}\uplus...\uplus Z_{n}=Z$.
In other words, each term of the sum considers a measurement-to-target
association hypothesis. Note that any hypothesis that assigns more
than one measurement to a target has zero likelihood, as indicated
in the last row of (\ref{eq:l_hat}). In the next example, we illustrate
how the sum in (\ref{eq:likelihood_standard}) is interpreted as it
is widely used in this paper.
\begin{example}
Let us consider $Z=\left\{ z_{1},z_{2}\right\} $ and $n=1$ so the
sum in (\ref{eq:likelihood_standard}) goes through all possible sets
$Z^{c}$ and $Z_{1}$ such that $Z^{c}\uplus Z_{1}=\left\{ z_{1},z_{2}\right\} $.
These are: 1) $Z^{c}=\emptyset$ and $Z_{1}=\left\{ z_{1},z_{2}\right\} $,
2) $Z^{c}=\left\{ z_{1}\right\} $ and $Z_{1}=\left\{ z_{2}\right\} $,
3) $Z^{c}=\left\{ z_{2}\right\} $ and $Z_{1}=\left\{ z_{1}\right\} $,
4) $Z^{c}=\left\{ z_{1},z_{2}\right\} $ and $Z_{1}=\emptyset$. Nevertheless,
as pointed out before, hypotheses that assign two measurements to
a target have probability zero so case 1) can be removed. 
\end{example}
\begin{center}
\begin{table}
\caption{\label{tab:Notations-for-different-likelihood}Notations in different
likelihood representations}
\begin{itemize}
\item $l\left(Z|X\right)$: Density of measurement set $Z$ given set $X$
of targets, defined in (\ref{eq:likelihood_standard}).
\item $\hat{l}\left(Z|x\right)$: Density of measurement set $Z$ given
target $x$, defined in (\ref{eq:l_hat}).
\item $\tilde{l}\left(z|Y\right)$: Likelihood of set $Y$ after observing
measurement $z$, defined in (\ref{eq:likelihood_Poisson2-1}).
\item $l_{o}\left(Z|Y,X_{1},...,X_{n}\right)$: Density of measurement set
$Z$ given sets $Y,X_{1},...,X_{n}$ $\left|X_{i}\right|\leq1$, defined
in (\ref{eq:likelihood_partitioning_conjugate_prior}).
\item $t\left(Z_{i}|X_{i}\right)$: Density of measurement $Z_{i}$ without
clutter given set $X_{i}$, $\left|X_{i}\right|\leq1$ , defined in
(\ref{eq:t_z_x}).
\end{itemize}
\end{table}
\par\end{center}

\section{Proof of the conjugacy of the PMBM\label{sec:Conjugate-priors-proof}}

In this section, we provide a non-PGFL proof of the conjugate prior
in \cite{Williams15b} for the standard point target measurement model.
We first review the conjugate prior in Section \ref{subsec:Conjugate-prior}.
Then, we proceed to derive the update for a Poisson prior in Section
\ref{subsec:Poisson_update}. Based on this preliminary derivation,
we perform a Bayesian update on the conjugate prior to show its conjugacy
in Section \ref{subsec:Update-of-conjugate}. The prediction step
is addressed in Section \ref{subsec:Prediction-conjugate-prior}.
We also establish the conjugacy property for multi-Bernoulli mixtures
in Section \ref{subsec:Conjugacy-MBM}.

\subsection{Conjugate prior\label{subsec:Conjugate-prior}}

It was proved in \cite{Williams15b} using PGFLs that the union of
two independent RFS, one Poisson and another a multi-Bernoulli mixture,
is conjugate with respect to the standard point target measurement
model. Before reviewing the mathematical form of the conjugate prior,
we give an overview of its key components and the underlying structure. 

\subsubsection{Interpretation}

The Poisson part of the conjugate prior models the undetected targets,
which represent targets that exist at the current time but have never
been detected. Each measurement at each time step gives rise to a
new potentially detected target. That is, there is the possibility
that a new measurement is the first detection of a target, but it
can also correspond to another previously detected target or clutter,
in which case there is no new target. As this target may exist or
not, its resulting distribution is Bernoulli and we refer to it as
``potentially detected target''.

In addition, for each potentially detected target, there are single
target association history hypotheses (single target hypotheses),
which represent possible histories of target-to-measurement (or misdetections)
associations. A single target hypothesis along with the existence
probability of the corresponding Bernoulli RFS incorporates information
about the events: the target never existed, the target exists at the
current time, the target did exist but death occurred at some point
since the last detection. Finally, a global association history hypothesis
(global hypothesis) contains one single target hypotheses for each
potential target with the constraints that each of the measurements
has to be contained in only one of the single target hypotheses.

\subsubsection{Mathematical representation}

Due to the independence property, the considered density is \cite{Mahler_book14}
\begin{align}
f\left(X\right) & =\sum_{Y\uplus W=X}f^{p}\left(Y\right)f^{mbm}\left(W\right)\label{eq:prior_prev}
\end{align}
where $f^{p}\left(\cdot\right)$ is a Poisson density and $f^{mbm}\left(\cdot\right)$
is a multi-Bernoulli mixture \cite{Williams15b}. The Poisson density
is
\begin{align}
f^{p}\left(X\right) & =e^{-\int\mu\left(x\right)dx}\left[\mu\left(\cdot\right)\right]^{X}\label{eq:Poisson_prior}
\end{align}
where $\mu\left(\cdot\right)$ represents its intensity. The multi-Bernoulli
mixture has multiplicative weights such that 
\begin{align}
f^{mbm}\left(X\right) & \propto\sum_{j}\sum_{X_{1}\uplus...\uplus X_{n}=X}\prod_{i=1}^{n}w_{j,i}f_{j,i}\left(X_{i}\right)\label{eq:multiBernoullimixture_multiplicative}
\end{align}
where $\propto$ stands for proportionality, $j$ is an index over
all global hypotheses (components of the mixtures) \cite{Williams15b},
$n$ is the number of potentially detected targets and, $w_{j,i}$
and $f_{j,i}\left(\cdot\right)$ are the weight and the Bernoulli
density of potentially detected target $i$ under the $j$th global
hypothesis. The Bernoulli densities have the expression
\begin{align}
f_{j,i}\left(X\right) & =\begin{cases}
1-r_{j,i} & X=\emptyset\\
r_{j,i}p_{j,i}\left(x\right) & X=\left\{ x\right\} \\
0 & \mathrm{otherwise}
\end{cases}\label{eq:Bernoulli_ji}
\end{align}
where $r_{j,i}$ is the probability of existence and $p_{j,i}\left(\cdot\right)$
is the state density given that it exists. Note that if there is only
one mixture component in the multi-Bernoulli mixture in (\ref{eq:multiBernoullimixture_multiplicative}),
i.e., $j$ can only take value $1$, we obtain a multi-Bernoulli density
\begin{align}
f^{mb}\left(X\right) & =\sum_{X_{1}\uplus...\uplus X_{n}=X}\prod_{i=1}^{n}f_{1,i}\left(X_{i}\right).\label{eq:MB_density}
\end{align}

The derivation demonstrates that a new Bernoulli component should
be created for each new measurement, where its existence corresponds
to the event that the measurement is the first detection of a new
target (which, prior to detection, was modelled by the Poisson component),
and non-existence corresponds to the event that the measurement is
a false alarm, or it corresponded to a different, previously detected
target. In addition, as each target can create at maximum one measurement,
the number of potentially detected targets corresponds to the number
of measurements up to the current time. The weight of global hypothesis
$j$ is proportional to the product of the hypothesis weights $\prod_{i=1}^{n}w_{j,i}$
for the $n$ potentially detected targets. If potentially detected
target $i$ is not considered in global hypothesis $j$, which implies
that its originating measurement was assigned to another target, $w_{j,i}=1$
and the probability of existence of $f_{j,i}\left(\cdot\right)$ is
zero. We do not make global hypotheses explicit in the notation as
it is not necessary to prove conjugacy. A notation that explicitly
states both these hypotheses and the data association history is provided
in \cite{Williams15b}.

Plugging (\ref{eq:multiBernoullimixture_multiplicative}) into (\ref{eq:prior_prev}),
we can also write (\ref{eq:prior_prev}) as 
\begin{align}
f\left(X\right) & \propto\sum_{Y\uplus X_{1}\uplus...\uplus X_{n}=X}f^{p}\left(Y\right)\sum_{j}\prod_{i=1}^{n}w_{j,i}f_{j,i}\left(X_{i}\right).\label{eq:prior}
\end{align}
Note that, given $X$, $X_{i}$ can be either empty or a single element
set (otherwise the density $f_{j,i}\left(\cdot\right)$ is zero) and
$Y$ can have any cardinality that meets the constraint $Y\uplus X_{1}\uplus...\uplus X_{n}=X$.

\subsection{Update of a Poisson prior\label{subsec:Poisson_update}}

In this section, we prove the update for a Poisson prior using the
likelihood (\ref{eq:likelihood_standard}). This result will be used
in Section \ref{subsec:Update-of-conjugate} to update the Poisson
component of the conjugate prior (\ref{eq:prior}).

\subsubsection{Likelihood representation}

\begin{figure}
\begin{centering}
\includegraphics[scale=0.8]{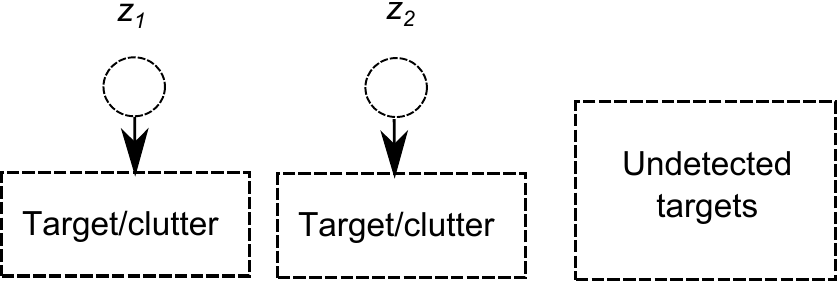}
\par\end{centering}
\caption{\label{fig:Likelihood_decomposition}Example of the likelihood decomposition
for $\left\{ z_{1},z_{2}\right\} $. Each measurement may have been
produced by a target or clutter. The likelihood also accounts for
the set of undetected targets. }
\end{figure}

For $Z=\left\{ z_{1},...,z_{m}\right\} $, we prove in Appendix \ref{sec:AppendixA}
that we can write the likelihood (\ref{eq:likelihood_standard}) as
\begin{align}
l\left(\left\{ z_{1},...,z_{m}\right\} |X\right) & =e^{-\lambda_{c}}\sum_{U\uplus Y_{1}\uplus...\uplus Y_{m}=X}\left[1-p_{d}\left(\cdot\right)\right]^{U}\nonumber \\
 & \quad\times\prod_{i=1}^{m}\tilde{l}\left(z_{i}|Y_{i}\right)\label{eq:likelihood_new_form}
\end{align}
where
\begin{align}
\tilde{l}\left(z|Y\right) & =\begin{cases}
p_{d}\left(y\right)p\left(z|y\right) & Y=\left\{ y\right\} \\
c\left(z\right) & Y=\emptyset\\
0 & \left|Y\right|>1.
\end{cases}\label{eq:likelihood_Poisson2-1}
\end{align}
The interpretation of (\ref{eq:likelihood_new_form}) is as follows.
We decompose the set $X$ of targets into all possible sets $U$,
$Y_{1}$,..., $Y_{m}$ such that $X=U\uplus Y_{1}...\uplus Y_{m}$.
Set $U$ represents the undetected targets and set $Y_{i}$ represents
the origin of the $i$th measurement, which can be a single-element
set containing the state of the target that gave rise to the measurement,
or an empty set if the measurement is clutter. This is a different
but equivalent way of expressing the data association hypotheses considered
in (\ref{eq:likelihood_standard}). An example is illustrated in Figure
\ref{fig:Likelihood_decomposition}. 

\subsubsection{Update}

Given a Poisson prior $f^{p}\left(\cdot\right)$ and $Z=\left\{ z_{1},...,z_{m}\right\} $,
we use Bayes' rule to compute the posterior $q^{p}\left(\cdot|Z\right)$
given the measurement set $Z$:
\begin{align}
q^{p}\left(X|Z\right) & \propto l\left(Z|X\right)f^{p}\left(X\right).\label{eq:BayesPoisson}
\end{align}
Note that $q^{p}\left(X|Z\right)$ denotes the updated Poisson process
with set $Z$ but this density is not Poisson unless $Z$ is empty.
We show in Appendix \ref{sec:AppendixB} that substituting (\ref{eq:Poisson_prior})
and (\ref{eq:likelihood_new_form}) into (\ref{eq:BayesPoisson}),
we find that the updated posterior is a union of a Poisson process
and a multi-Bernoulli RFS such that
\begin{align}
 & q^{p}\left(X|Z\right)\nonumber \\
 & \propto\sum_{U\uplus Y_{1}\uplus...\uplus Y_{m}=X}q^{p}\left(U\right)\prod_{i=1}^{m}\rho^{p}\left(z_{i}\right)q^{p}\left(Y_{i}|z_{i}\right)\label{eq:update_Poisson1}\\
 & \propto\sum_{U\uplus Y_{1}\uplus...\uplus Y_{m}=X}q^{p}\left(U\right)\prod_{i=1}^{m}q^{p}\left(Y_{i}|z_{i}\right)\label{eq:update_Poisson2}
\end{align}
where the Poisson component has the intensity of the prior multiplied
by $\left(1-p_{d}\left(\cdot\right)\right)$ 
\begin{align}
q^{p}\left(U\right) & \propto\left[\left(1-p_{d}\left(\cdot\right)\right)\mu\left(\cdot\right)\right]^{U}\label{eq:undetected_targets}
\end{align}
and the Bernoulli components are given by 
\begin{align}
q^{p}\left(Y_{i}|z_{i}\right) & =\tilde{l}\left(z_{i}|Y_{i}\right)f^{p}\left(Y_{i}\right)/\left(e^{-\int\mu\left(x\right)dx}\rho^{p}\left(z_{i}\right)\right)\label{eq:Bernoulli_Poisson_component1}\\
 & =\begin{cases}
1-r^{p}\left(z_{i}\right) & Y_{i}=\emptyset\\
r^{p}\left(z_{i}\right)p^{p}\left(y|z_{i}\right) & Y_{i}=\left\{ y\right\} \\
0 & \mathrm{otherwise}
\end{cases}\label{Bernoulli_Poisson_component1_evaluation}
\end{align}
where
\begin{align}
\rho^{p}\left(z_{i}\right) & =\int\tilde{l}\left(z_{i}|Y_{i}\right)f^{p}\left(Y_{i}\right)\delta Y_{i}/e^{-\int\mu\left(x\right)dx}\nonumber \\
 & =c\left(z_{i}\right)+e\left(z_{i}\right)\label{eq:ro_Bernoulli_Poisson}\\
e\left(z_{i}\right) & =\int p\left(z_{i}|y\right)p_{d}\left(y\right)\mu\left(y\right)dy\\
r^{p}\left(z_{i}\right) & =e\left(z_{i}\right)/\rho^{p}\left(z_{i}\right)\label{eq:Bernoulli_Poisson_prob_existence}\\
p^{p}\left(y|z_{i}\right) & =p_{d}\left(y\right)p\left(z_{i}|y\right)\mu\left(y\right)/e\left(z_{i}\right).\label{eq:Bernoulli_Poisson_last}
\end{align}
 Note that we define $\rho^{p}\left(z_{i}\right)$ by normalising
it by $e^{-\int\mu\left(x\right)dx}$ as (\ref{eq:ro_Bernoulli_Poisson})
will be used later on and there is no need to compute this exponential
in the resulting filter.

The explanation of the resulting updated density (\ref{eq:update_Poisson2})
is as follows. Given $Z=\left\{ z_{1},...,z_{m}\right\} $ and a Poisson
process with intensity $\mu\left(\cdot\right)$, the updated density
is the union of $m+1$ independent random finite sets, represented
by $U,Y_{1},...,Y_{m}$. RFS $U$ is Poisson with intensity $\left(1-p_{d}\left(\cdot\right)\right)\mu\left(\cdot\right)$
and represents the undetected part of the prior. RFS $Y_{j}$ is the
Bernoulli RFS coming from the $j$th measurement. Its density is given
by (\ref{eq:Bernoulli_Poisson_component1}), which has a probability
of existence given by (\ref{eq:Bernoulli_Poisson_prob_existence}).

\subsection{Update of conjugate prior\label{subsec:Update-of-conjugate}}

In order to show the update of the conjugate prior, we first propose
another likelihood representation in Section \ref{subsec:Likelihood-representation}.
Then, we show the update of one Bernoulli component in Section \ref{subsec:Update-of-one-Bernoulli}
and utilise this result to obtain the whole update in Section \ref{subsec:Update-conjugate-prior2}. 

\subsubsection{Likelihood representation\label{subsec:Likelihood-representation}}

Here we represent the likelihood in a way that is suitable to update
the Poisson multi-Bernoulli mixture. For any sets $Y,X_{1},...,X_{n}$
such that $\left|X_{i}\right|\leq1$ for $i=1,...,n$ we define the
function
\begin{align}
l_{o}\left(Z|Y,X_{1},...,X_{n}\right) & =\sum_{Z_{1}\uplus...\uplus Z_{n}\uplus Z^{y}=Z}l\left(Z^{y}|Y\right)\nonumber \\
 & \quad\times\prod_{i=1}^{n}t\left(Z_{i}|X_{i}\right).\label{eq:likelihood_partitioning_conjugate_prior}
\end{align}
where $Z^{y}$ represents both measurements from targets in $Y$ and
clutter, and $t\left(Z_{i}|X_{i}\right)$ is the likelihood for a
set with zero or one measurement elements without clutter
\begin{align}
t\left(Z_{i}|X_{i}\right) & =\begin{cases}
p_{d}\left(x\right)l\left(z|x\right) & Z_{i}=\left\{ z\right\} ,X_{i}=\left\{ x\right\} \\
1-p_{d}\left(x\right) & Z_{i}=\emptyset,X_{i}=\left\{ x\right\} \\
1 & Z_{i}=\emptyset,X_{i}=\emptyset\\
0 & \mathrm{otherwise}.
\end{cases}\label{eq:t_z_x}
\end{align}
We show in Appendix \ref{sec:AppendixC} that for any sets $Y,X_{1},...,X_{n}$,
such that $\left|X_{i}\right|\leq1$ for $i=1,...,n$, we have
\begin{align}
l_{o}\left(Z|Y,X_{1},...,X_{n}\right) & =l\left(Z|X\right)\label{eq:likelihood_equivalence_proof}
\end{align}
where $X=Y\uplus X_{1}\uplus...\uplus X_{n}$. That is, the evaluation
of function $l_{o}\left(Z|\cdot,\cdot,...,\cdot\right)$ at any sets
$Y,X_{1},...,X_{n}$, such that $\left|X_{i}\right|\leq1$ for $i=1,...,n$,
is equivalent to the evaluation of the likelihood $l\left(Z|\cdot\right)$
at set $X=Y\uplus X_{1}\uplus...\uplus X_{n}$. 

\subsubsection{Update of one Bernoulli component\label{subsec:Update-of-one-Bernoulli}}

As will be seen in the next subsection, one part of the update of
the conjugate prior requires the update of the Bernoulli components.
Therefore, we proceed to derive this update in this subsection so
that we have the result available for the next subsection. In the
update of the conjugate prior, we will need to compute the update
of Bernoulli component $f_{j,i}\left(\cdot\right)$, which is given
by (\ref{eq:Bernoulli_ji}), by measurement $Z_{i}$ considering the
likelihood $t\left(Z_{i}|\cdot\right)$. We denote the corresponding
updated density as
\begin{align}
q_{j,i}\left(X_{i}|Z_{i}\right) & =t\left(Z_{i}|X_{i}\right)f_{j,i}\left(X_{i}\right)/\rho_{j,i}\left(Z_{i}\right)\label{eq:update_Bernoulli_component}
\end{align}
where the numerator is the joint density of $Z_{i}$ and $X_{i}$
and 
\begin{align}
\rho_{j,i}\left(Z_{i}\right) & =\int t\left(Z_{i}|X\right)f_{j,i}\left(X\right)\delta X.\label{eq:ro_old_Bernoulli_component}
\end{align}

According to $t\left(Z_{i}|X\right)$ in (\ref{eq:t_z_x}), $Z_{i}$
can only take values $Z_{i}=\left\{ z\right\} $ or $Z_{i}=\emptyset$
so that the likelihood is different from zero so we proceed to compute
(\ref{eq:update_Bernoulli_component}) in these two cases. For $Z_{i}=\left\{ z\right\} $,
$t\left(Z_{i}|X\right)$ is only different from zero if $X=\left\{ x\right\} $
so, using (\ref{eq:ro_old_Bernoulli_component}), (\ref{eq:t_z_x})
and (\ref{eq:Bernoulli_ji}), we obtain
\begin{align}
\rho_{j,i}\left(\left\{ z\right\} \right) & =r_{j,i}\int p_{d}\left(x\right)l\left(z|x\right)p_{j,i}\left(x\right)dx.\label{eq:ro_j_i_empty}
\end{align}
Substituting the previous equations into (\ref{eq:update_Bernoulli_component})
we find that $q_{j,i}\left(\cdot|\left\{ z\right\} \right)$ is Bernoulli
with probability of existence 1 and target state density proportional
to $p_{d}\left(x\right)l\left(z|x\right)p_{j,i}\left(x\right)$. For
$Z_{i}=\emptyset$, $t\left(Z_{i}|X\right)$ can be different from
zero if $X=\left\{ x\right\} $ or $X=\emptyset$. Now, using (\ref{eq:ro_old_Bernoulli_component}),
(\ref{eq:t_z_x}) and (\ref{eq:Bernoulli_ji}), we have 
\begin{align}
\rho_{j,i}\left(\emptyset\right) & =1-r_{j,i}+r_{j,i}\int\left(1-p_{d}\left(x\right)\right)p_{j,i}\left(x\right)dx.\label{eq:ro_j_i_one_element}
\end{align}
Then, substituting the previous equations into (\ref{eq:update_Bernoulli_component}),
we find that $q_{j,i}\left(\cdot|\emptyset\right)$ is Bernoulli with
probability of existence
\begin{align*}
 & r_{j,i}\left[\int\left(1-p_{d}\left(x\right)\right)p_{j,i}\left(x\right)dx\right]/\rho_{j,i}\left(\emptyset\right)
\end{align*}
and target state density proportional to $\left(1-p_{d}\left(x\right)\right)p_{j,i}\left(x\right)$. 

\subsubsection{Update of the conjugate prior\label{subsec:Update-conjugate-prior2}}

Substituting the prior (\ref{eq:prior}) into Bayes' rule (\ref{eq:Bayes_rule}),
we have that
\begin{align*}
 & q\left(X|Z\right)\\
 & \propto\sum_{Y\uplus X_{1}\uplus...\uplus X_{n}=X}l\left(Z|X\right)f^{p}\left(Y\right)\sum_{j}\prod_{i=1}^{n}w_{j,i}f_{j,i}\left(X_{i}\right)\\
 & =\sum_{Y\uplus X_{1}\uplus...\uplus X_{n}=X}l\left(Z|Y\uplus X_{1}\uplus...\uplus X_{n}\right)f^{p}\left(Y\right)\\
 & \quad\times\sum_{j}\prod_{i=1}^{n}w_{j,i}f_{j,i}\left(X_{i}\right).
\end{align*}
As $f_{j,i}\left(\cdot\right)$ is Bernoulli, the corresponding term
in the previous sum is different from zero if and only if $\left|X_{i}\right|\leq1$.
Therefore, we can add this constraint to the sum:
\begin{align}
 & q\left(X|Z\right)\nonumber \\
 & \propto\sum_{Y\uplus X_{1}\uplus...\uplus X_{n}=X:\left|X_{i}\right|\leq1,\forall i}l\left(Z|Y\uplus X_{1}\uplus...\uplus X_{n}\right)f^{p}\left(Y\right)\nonumber \\
 & \quad\times\sum_{j}\prod_{i=1}^{n}w_{j,i}f_{j,i}\left(X_{i}\right).\label{eq:update_conjugate_prior1a}
\end{align}
Now, substitute (\ref{eq:likelihood_equivalence_proof}) in (\ref{eq:update_conjugate_prior1a})
so that 
\begin{align}
 & q\left(X|Z\right)\nonumber \\
 & \propto\sum_{Y\uplus X_{1}\uplus...\uplus X_{n}=X:\left|X_{i}\right|\leq1,\forall i}l_{o}\left(Z|Y,X_{1},...,X_{n}\right)f^{p}\left(Y\right)\nonumber \\
 & \quad\times\sum_{j}\prod_{i=1}^{n}w_{j,i}f_{j,i}\left(X_{i}\right).\nonumber \\
 & =\sum_{Y\uplus X_{1}\uplus...\uplus X_{n}=X}\sum_{Z=Z_{1}\uplus...\uplus Z_{n}\uplus Z^{y}}\left[l\left(Z^{y}|Y\right)f^{p}\left(Y\right)\right]\nonumber \\
 & \quad\times\sum_{j}\left[\prod_{i=1}^{n}w_{j,i}t\left(Z_{i}|X_{i}\right)f_{j,i}\left(X_{i}\right)\right].\label{eq:update_conjugate_prior1b}
\end{align}
Factor $l\left(Z^{y}|Y\right)f^{p}\left(Y\right)$ in (\ref{eq:update_conjugate_prior1b})
represents the unnormalised update of a Poisson prior. In (\ref{eq:update_Poisson1}),
we obtained the result for such an update so we can apply it in (\ref{eq:update_conjugate_prior1b}).
Therefore, we have that 
\begin{align}
 & q\left(X|Z\right)\nonumber \\
 & \propto\sum_{Y\uplus X_{1}\uplus...\uplus X_{n}=X}\sum_{Z=Z_{1}\uplus...\uplus Z_{n}\uplus Z^{y}}\sum_{U\uplus Y_{1}...\uplus Y_{m}=Y}q^{p}\left(U\right)\nonumber \\
 & \:\times\prod_{i=1}^{m}\left[\chi_{Z^{y}}\left(z_{i}\right)\rho^{p}\left(z_{i}\right)q^{p}\left(Y_{i}|z_{i}\right)+\left(1-\chi_{Z^{y}}\left(z_{i}\right)\right)\delta_{\emptyset}\left(Y_{i}\right)\right]\nonumber \\
 & \:\times\sum_{j}\left[\prod_{i=1}^{n}w_{j,i}t\left(Z_{i}|X_{i}\right)f_{j,i}\left(X_{i}\right)\right]\label{eq:update_conjugate_prior1}
\end{align}
where $\chi_{A}\left(\cdot\right)$ denotes the indicator function
on set $A$ 
\begin{align*}
\chi_{A}\left(z\right) & =\begin{cases}
0 & z\notin A\\
1 & z\in A
\end{cases}
\end{align*}
and $\delta_{\emptyset}\left(\cdot\right)$ is the multi-target Dirac
delta centered at $\emptyset$ \cite[Eq. (11.124)]{Mahler_book07}:
\begin{align*}
\delta_{\emptyset}\left(Y\right) & =\begin{cases}
0 & Y\neq\emptyset\\
1 & Y=\emptyset.
\end{cases}
\end{align*}
We should note that for the update of the Poisson RFS $Y$, we only
consider the measurements that are hypothesised to be coming from
$Y$, which are represented by $Z^{y}$ in (\ref{eq:update_conjugate_prior1}).
Therefore, in the third line of (\ref{eq:update_conjugate_prior1}),
we use a product over measurements $z_{1},...,z_{m}$ but setting
the probability of existence of the Bernoulli RFS associated to $z_{i}$
to zero if $z_{i}$ is not included in $Z^{y}$, $\chi_{Z^{y}}\left(z_{i}\right)=0$.

Simplifying (\ref{eq:update_conjugate_prior1}), we have
\begin{align}
 & q\left(X|Z\right)\nonumber \\
 & \propto\sum_{U\uplus X_{1}\uplus...\uplus X_{n}\uplus Y_{1}\uplus...\uplus Y_{m}=X}q^{p}\left(U\right)\sum_{j}\sum_{Z_{1}\uplus...\uplus Z_{n}\uplus Z^{y}=Z}\nonumber \\
 & \:\times\prod_{i=1}^{m}\left[\chi_{Z^{y}}\left(z_{i}\right)\rho^{p}\left(z_{i}\right)q\left(Y_{i}|z_{i}\right)+\left(1-\chi_{Z^{y}}\left(z_{i}\right)\right)\delta_{\emptyset}\left(Y_{i}\right)\right]\nonumber \\
 & \:\times\left[\prod_{i=1}^{n}w_{j,i}\rho_{j,i}\left(Z_{i}\right)q_{j,i}\left(X_{i}|Z_{i}\right)\right].\label{eq:update_conjugate_prior2}
\end{align}

Merging the two inner summations into one, rearranging the indices
and comparing with the prior (\ref{eq:prior}), we see that the posterior
is also the union of two independent processes: one Poisson and the
other a multi-Bernoulli mixture. This proves that this density is
conjugate with respect to the standard point target measurement model. 

We would also like to comment on the weights of the new potentially
detected targets, which are considered in the product over $m$ factors
in (\ref{eq:update_conjugate_prior2}). If a new potentially detected
target $i$ does not exist in a new global hypothesis, which implies
that $\chi_{Z^{y}}\left(z_{i}\right)=0$, then, its hypothesis weight
is one and its density $\delta_{\emptyset}\left(Y_{i}\right)$ can
also be represented as Bernoulli with zero probability of existence.
On the contrary, if a new potentially detected target $i$ exists
in a new global hypothesis, $\chi_{Z^{y}}\left(z_{i}\right)=1$, its
hypothesis weight is $\rho^{p}\left(z_{i}\right)$ and its Bernoulli
density is given by $q\left(Y_{i}|z_{i}\right)$. The weight for a
previous potentially detected target corresponds to the same weight
$w_{j,i}$ multiplied by $\rho_{j,i}\left(Z_{i}\right)$, see (\ref{eq:ro_old_Bernoulli_component}).
Depending on the hypothesis $Z_{i}$ can be either empty or has one
element, the resulting weights and Bernoulli components in these two
cases are discussed after (\ref{eq:ro_old_Bernoulli_component}).

\subsection{Prediction of the conjugate prior\label{subsec:Prediction-conjugate-prior}}

In this section, we prove that, if the posterior is a PMBM of the
form (\ref{eq:prior_prev})-(\ref{eq:multiBernoullimixture_multiplicative}),
then the prior at the next time step is also PMBM with the following
parameters. The Poisson part of the predicted density is obtained
using the PHD filter prediction equation \cite{Mahler03} so that
its intensity is 
\begin{align*}
\mu\left(x\right) & =\lambda^{b}\left(x\right)+\int g\left(x|y\right)p_{s}\left(y\right)\lambda^{u}\left(y\right)dy
\end{align*}
where $\lambda^{u}\left(\cdot\right)$ denotes the intensity of the
Poisson part of the posterior. In addition, if the parameters of the
posterior multi-Bernoulli mixture are $w_{j,i}^{u}$, $p_{j,i}^{u}\left(\cdot\right)$,
$r_{j,i}^{u}$, the predicted parameters are given by the multi-target
multi-Bernoulli (MeMBer) filter prediction equation \cite{Mahler07}
\begin{align*}
w_{j,i} & =w_{j,i}^{u}\\
r_{j,i} & =r_{j,i}^{u}\int p_{j,i}^{u}\left(y\right)p_{s}\left(y\right)dy\\
p_{j,i}\left(x\right) & \propto\int g\left(x|y\right)p_{s}\left(y\right)p_{j,i}^{u}\left(y\right)dy.
\end{align*}

In order to prove this result, we first note the equivalences between
the dynamic/measurement processes \cite[Chap. 13]{Mahler_book07}.
In the standard models, each target is detected/survives with probability
$p_{d}\left(\cdot\right)/p_{s}\left(\cdot\right)$ and generates a
measurement/new target state according to $l\left(\cdot|\cdot\right)/g\left(\cdot|\cdot\right)$
and there are additional independent clutter measurements/new born
targets distributed according to a Poisson process with intensity
$c\left(\cdot\right)/\lambda^{b}\left(\cdot\right)$. In other words,
the density of the measurement, denoted as $\rho(\cdot)$ in (\ref{eq:PDF_measurement}),
is equivalent to the predicted density, denoted as $\omega\left(\cdot\right)$
in (\ref{eq:prediction_step}), by making the previous equivalences
\cite{Angel15_d}. As we have explained the notation for proving the
update step, we will first compute the density of the measurements
and then establish the equivalence with the prediction step. Before
doing so, we establish the following corollary.
\begin{cor}
\label{cor:Independent_set_integralr}Let us consider an RFS $X=X_{1}\uplus...\uplus X_{n}$
where $X_{1},...,$$X_{n}$ are independent so the density $f\left(\cdot\right)$
of $X$ can be written as
\begin{align*}
f\left(X\right) & =\sum_{X_{1}\uplus...\uplus X_{n}=X}\prod_{i=1}^{n}f_{i}\left(X_{i}\right)
\end{align*}
where $f_{i}\left(\cdot\right)$ is the density of $X_{i}$. For an
arbitrary set-valued function $v\left(\cdot\right)$, then
\begin{align*}
 & \int v\left(X\right)f\left(X\right)\delta X\\
 & \quad=\int...\int v\left(X_{1}\cup...\cup X_{n}\right)\prod_{i=1}^{n}f_{i}\left(X_{i}\right)\delta X_{1}...\delta X_{n}.
\end{align*}
\end{cor}
The proof of the corollary is straightforward using \cite[Eq. (63)]{Williams15}
$n-1$ times. Substituting (\ref{eq:prior}) into (\ref{eq:PDF_measurement}),
we obtain
\begin{align*}
\rho\left(Z\right) & \propto\sum_{j}\left[\prod_{i=1}^{n}w_{j,i}\right]\int l(Z|X)\\
 & \quad\times\sum_{Y\uplus X_{1}\uplus...\uplus X_{n}=X}f^{p}\left(Y\right)\prod_{i=1}^{n}f_{j,i}\left(X_{i}\right)\delta X.
\end{align*}
where $l(\cdot|X)$ is the density of the measurements (including
clutter) given $X$. Using Corollary \ref{cor:Independent_set_integralr},
we find
\begin{align*}
\rho\left(Z\right) & \propto\sum_{j}\left[\prod_{i=1}^{n}w_{j,i}\right]\int\int...\int l(Z|Y\cup X_{1}\cup...\cup X_{n})\\
 & \quad\times f^{p}\left(Y\right)\prod_{i=1}^{n}f_{j,i}\left(X_{i}\right)\delta Y\delta X_{1}...\delta X_{n}.
\end{align*}
As $f_{j,i}\left(\cdot\right)$ are Bernoulli, we can apply (\ref{eq:likelihood_equivalence_proof})
and then (\ref{eq:likelihood_partitioning_conjugate_prior}) so that
\begin{align*}
\rho\left(Z\right) & \propto\sum_{j}\left[\prod_{i=1}^{n}w_{j,i}\right]\int\int...\int l_{o}\left(Z|Y,X_{1},...,X_{n}\right)\\
 & \quad\times f^{p}\left(Y\right)\prod_{i=1}^{n}f_{j,i}\left(X_{i}\right)\delta Y\delta X_{1}...\delta X_{n}\\
 & =\sum_{j}\sum_{Z_{1}\uplus...\uplus Z_{n}\uplus Z^{y}=Z}\int l\left(Z^{y}|Y\right)f^{p}\left(Y\right)\delta Y\\
 & \quad\times\left[\prod_{i=1}^{n}w_{j,i}\int t\left(Z_{i}|X_{i}\right)f_{j,i}\left(X_{i}\right)\delta X_{i}\right]\\
 & =\sum_{j}\sum_{Z_{1}\uplus...\uplus Z_{n}\uplus Z^{y}=Z}\int l\left(Z^{y}|Y\right)f^{p}\left(Y\right)\delta Y\\
 & \quad\times\left[\prod_{i=1}^{n}w_{j,i}\rho_{j,i}\left(Z_{i}\right)\right]
\end{align*}
where we recall that $\rho_{j,i}\left(\cdot\right)$ is a Bernoulli
density previously specified in (\ref{eq:ro_j_i_empty}) and (\ref{eq:ro_j_i_one_element})
and $t\left(\cdot|X\right)$ is the density of the measurement generated
by a set $X$, which can have cardinality zero or one, without clutter.
From the PHD filter recursion \cite{Mahler03,Angel15_d}, we know
that $\int l\left(Z^{y}|Y\right)f^{p}\left(Y\right)\delta Y$ is a
Poisson density on $Z^{y}$ with intensity $c\left(x\right)+\int p\left(x|y\right)p_{d}\left(y\right)\mu\left(y\right)dy$. 

In summary, the density of the measurement is the union of a Poisson
process and a multi-Bernoulli mixture with the same weights as the
prior and the parameters specified above. Due to the equivalence of
parameters in the prediction/update steps mentioned at the beginning
of this section, the proof of the conjugacy of the PMBM is finished. 

\subsection{Conjugacy for multi-Bernoulli mixtures\label{subsec:Conjugacy-MBM}}

In this section, we establish the conjugacy property of multi-Bernoulli
mixtures (MBM), which results in the MBM filter. This result will
help us establish relations between PMBM and labelled conjugate priors,
see Section \ref{sec:Connection-delta-GLMB}.
\begin{cor}
\label{cor:MBM_conjugacy}If the birth process is multi-Bernoulli
or MBM, the family of MBM is a conjugate prior for the standard point
target measurement and dynamic models.
\end{cor}
The update step can be performed as above by setting the intensity
of the Poisson density to zero and the prediction step is proved in
Appendix \ref{sec:AppendixC_2_multiBernoulli_mixture_birth}. In the
prediction step, for multi-Bernoulli birth, we incorporate additional
multi-Bernoulli components to each term in the mixture. For multi-Bernoulli
mixture birth, a new term is created for each combination of a term
in the old mixture and a term in the birth mixture, where the new
term combines the Bernoulli components from each. 

\section{Connection between the PMBM filter and the $\delta$-GLMB filter\label{sec:Connection-delta-GLMB}}

In this section, we establish the connection between the PMBM filter
and the $\delta$-GLMB filter. In order to do so, we first discuss
an alternative parameterisation of multi-Bernoulli mixtures in Section
\ref{subsec:MBM01}. Then, we introduce the conjugacy properties of
labelled MBMs in Section \ref{subsec:Conjugacy-of-labelledMBM}. Section
\ref{subsec:-GLMB-density-isLMBM} proves that the $\delta$-GLMB
density is in fact a labelled multi-Bernoulli mixture, but with a
less efficient parameterisation from a storage and computational point
of view. A discussion on both parameterisations and the advantages
of the PMBM form is given in Section \ref{subsec:Discussion}.

\subsection{Multi-Bernoulli mixture 01 parameterisation\label{subsec:MBM01}}

In this subsection, we explain the MBM$_{01}$ parameterisation, which
is an alternative parameterisation of an MBM in which the Bernoulli
densities have existence probabilities that are either zero or one.
The MBM$_{01}$ parameterisation is relevant to the connection between
the PMBM filter and the $\delta$-GLMB filter, as will be explained
in the following subsections. The MBM parameterisation in (\ref{eq:multiBernoullimixture_multiplicative})
is simply referred to as the MBM parameterisation.

We first explain the MBM$_{01}$ parameterisation of a single Bernoulli
density. A Bernoulli density $f_{j,i}\left(\cdot\right)$, see (\ref{eq:Bernoulli_ji}),
can be written as a mixture of Bernoulli densities with existence
probabilities that are either zero or one as
\begin{align}
f_{j,i}\left(X_{i}\right) & =\left(1-r_{j,i}\right)f_{j,i}^{0}\left(X_{i}\right)+r_{j,i}f_{j,i}^{1}\left(X_{i}\right)\label{eq:Bernoulli_mixture}
\end{align}
where 
\begin{align}
f_{j,i}^{\theta_{i}}\left(X_{i}\right) & =\begin{cases}
1-\theta_{i} & X_{i}=\emptyset\\
\theta_{i}p_{j,i}\left(x\right) & X_{i}=\left\{ x\right\} \\
0 & \mathrm{otherwise}
\end{cases}\label{eq:Bernoulli_ji_01}
\end{align}
for $\theta_{i}\in\left\{ 0,1\right\} $. It should be noted that
if $r_{j,i}\in(0,1)$, the mixture in (\ref{eq:Bernoulli_mixture})
has two components, otherwise, it has one component. We say that $f_{j,i}^{0}\left(\cdot\right)$
and $f_{j,i}^{1}\left(\cdot\right)$ have deterministic existence,
since $X_{i}=\emptyset$ and $\left|X_{i}\right|=1$ have probability
one for $f_{j,i}^{0}\left(\cdot\right)$ and $f_{j,i}^{1}\left(\cdot\right)$,
respectively. 

In an MBM, we can expand all Bernoulli densities in a similar way,
such that existence probabilities of all Bernoulli densities are either
0 or 1. For instance, the MBM in (\ref{eq:multiBernoullimixture_multiplicative})
can be written in MBM$_{01}$ parameterisation as 
\begin{align}
f^{mbm}\left(X\right) & \propto\sum_{j}\sum_{\theta\in\left\{ 0,1\right\} ^{n}}\sum_{X_{1}\uplus...\uplus X_{n}=X}\prod_{i=1}^{n}w_{j,i}v_{j,i,\theta_{i}}f_{j,i}^{\theta_{i}}\left(X_{i}\right),\label{eq:MBM_01_parameterisation}
\end{align}
where $\theta=\left(\theta_{1},...,\theta_{n}\right)$, $v_{j,i,\theta_{i}}=\left(1-r_{j,i}\right)^{1-\theta_{i}}r_{j,i}^{\theta_{i}}$
and $\left\{ 0,1\right\} ^{n}$ represents $n$ Cartesian products
of $\left\{ 0,1\right\} $. From (\ref{eq:MBM_01_parameterisation}),
we can directly establish the following proposition. 
\begin{prop}
\label{prop:MBM01_number_components}Consider an MBM with $m$ mixture
components. Let $n_{j}$ denote the number of Bernoulli densities,
in component $j$ of the MBM, with existence probability in the interval
$(0,1)$. Then, the MBM$_{01}$ parameterisation of the MBM requires
$\sum{}_{j=1}^{m}2^{n_{j}}$ mixture components.
\end{prop}
Let us illustrate the increase in the number of mixture components
(global hypotheses) with the following example.
\begin{example}
\label{exa:MBM01}Consider an MB density (MBM with one mixture component)
with three targets and existence probabilities $r_{1,1}=0.8$, $r_{1,2}=0.2$
and $r_{1,3}=1$. The corresponding MBM$_{01}$ parameterisation contains
4 mixture components (global hypotheses) with weights $r_{1,1}r_{1,2}$,
$\left(1-r_{1,1}\right)r_{1,2}$, $r_{1,1}\left(1-r_{1,2}\right)$
and $\left(1-r_{1,1}\right)\left(1-r_{1,2}\right)$.
\end{example}
It should be noted that, according to Proposition \ref{prop:MBM01_number_components},
the MBM$_{01}$ parameterisation can give rise to a tremendous increase
in the number of components in the mixture (global hypotheses), which
is an inefficient way to represent an MBM distribution. In fact, we
can use the PMBM filter with an MBM$_{01}$ parameterization, but
a standard brute-force implementation would yield much higher computational
complexity due to the increase in the number of global hypotheses.
For instance, as will be clarified in Section \ref{sec:Linear-Gaussian-implementation},
we need to solve a data-association problem for each global hypothesis
so it is desirable to have as few global hypotheses as possible.

\subsection{Conjugacy of labelled multi-Bernoulli mixtures\label{subsec:Conjugacy-of-labelledMBM}}

In this section, we prove the conjugacy for labelled multi-Bernoulli
mixtures. In the labelled approach, we augment the single target state
space with a label, which is a variable that is unique for each new
born target and fixed with time \cite{Angel13,Vo13}. A labelled MBM
is therefore obtained by adding (unique) labels to an MBM, see (\ref{eq:multiBernoullimixture_multiplicative}),
which results in a density of the form
\begin{align}
f\left(X\right) & \propto\sum_{j}\sum_{X_{1}\uplus...\uplus X_{n}=X}\prod_{i=1}^{n}w_{j,i}f_{j,i}^{lb}\left(X_{i}\right)\label{eq:labelled_MB_mixture}
\end{align}
where $f_{j,i}^{lb}\left(\cdot\right)$ is the labelled Bernoulli
density for target $i$ for mixture component $j$ given by 
\begin{align}
f_{j,i}^{lb}\left(X\right) & =\begin{cases}
1-r_{j,i} & X=\emptyset\\
r_{j,i}p_{j,i}\left(x\right)\delta\left[\ell-\ell_{i}\right] & X=\left\{ \left(x,\ell\right)\right\} \\
0 & \mathrm{otherwise}.
\end{cases}\label{eq:labelled_Bernoulli}
\end{align}
Here, $\delta\left[\cdot\right]$ represents a Kronecker delta, $\ell_{i}$
is the deterministic label of target $i$, and $r_{j,i}$ and $p_{j,i}\left(\cdot\right)$
are its existence probability and state density for global hypothesis
$j$. In addition, in (\ref{eq:labelled_MB_mixture}), we have $\ell_{i}\neq\ell_{i'}$
for $i\neq i'$ to ensure unique labels. The main difference between
(\ref{eq:labelled_Bernoulli}) and its unlabelled counterpart (\ref{eq:Bernoulli_ji})
is that the state space has been expanded to incorporate a unique
label that is known for each $i$. Note that the labelled MBM in (\ref{eq:labelled_MB_mixture})
can also be written in (labelled) MBM$_{01}$ parameterisation analogously
to how (\ref{eq:multiBernoullimixture_multiplicative}) was expressed
in (\ref{eq:MBM_01_parameterisation}).

We establish the following corollary. 
\begin{cor}
\label{cor:Labelled_MBM_conjugacy}If the birth process is labelled
multi-Bernoulli or labelled MBM, whose targets have unique labels,
and labels are fixed with time, the family of labelled MBM is a conjugate
prior for the standard point target measurement and dynamic models.
\end{cor}
As we explain in this paragraph, Corollary \ref{cor:Labelled_MBM_conjugacy}
is a particular case of Corollary \ref{cor:MBM_conjugacy} by considering
the specific properties of the labels: they are unique and fixed with
time. Note that, in this paper, we have denoted the single target
state as $x$, without any assumptions on it so it is flexible enough
to include a label, without specifying it explicitly. In order to
prove conjugacy for labelled MBM, we just need to model that one component
of the target state (the label) is unique and fixed  using the general
birth/dynamic models. This is done by considering labelled MB or MBM
birth process and a single target transition density $g\left(\cdot|\cdot\right)$
that has the constraint that the label does not change with time.
 Therefore, the conjugacy for labelled MBM is just a particular case
of MBM conjugacy, with the previous constraints in the birth model
and single transition density. As a result, the prediction and update
equations for the general MBM filter are also valid for the labelled
MBM filter.

\subsection{Relation between $\delta$-GLMB densities and labelled multi-Bernoulli
mixtures\label{subsec:-GLMB-density-isLMBM}}

The most common conjugate prior for labeled RFSs is the $\delta$-GLMB
density \cite{Vo13}, and in the following proposition, which is proved
in Appendix \ref{sec:AppendixD_deltaGLMB}, we relate a $\delta$-GLMB
density to a labelled MBM density.
\begin{prop}
\label{prop:GLMB_density_equivalence} $\delta$-GLMB and labelled
MBM with MBM$_{01}$ parameterisation can represent the same labelled
multi-target densities with the same number of global hypotheses,
in which target existence is deterministic.
\end{prop}
As indicated in the previous proposition, $\delta$-GLMB and labelled
MBM with MBM$_{01}$ parameterisations have the same type of global
hypotheses, in the sense that both consider global hypotheses with
deterministic target existence and labelled targets. One difference,
however, is that the $\delta$-GLMB notation \cite{Vo13,Vo14} can
only consider labelled targets, while the MBM$_{01}$ notation can
handle labelled and unlabelled targets. According to Proposition \ref{prop:GLMB_density_equivalence},
the number of global hypotheses (mixture components) in the $\delta$-GLMB
density in relation to a (labelled) MBM parameterisation is the same
as in the (labelled) MBM$_{01}$ parameterisation, which is given
by Proposition \ref{prop:MBM01_number_components}. This is illustrated
in the next example.
\begin{example}
Suppose distinct labels $\ell_{1}$, $\ell_{2}$, $\ell_{3}$ are
added to the three Bernoulli components in Example \ref{exa:MBM01},
such that we have a labelled MB density (labelled MBM with one mixture
component). As in Example \ref{exa:MBM01}, its MBM$_{01}$/ $\delta$-GLMB
parameterisations have four mixture components (global hypotheses),
with the same weights as in Example \ref{exa:MBM01}. 
\end{example}

\subsection{Discussion\label{subsec:Discussion}}

We proceed to discuss some computational and implementational advantages
of the MBM parameterisation (either labelled or not) compared to the
MBM$_{01}$ and $\delta$-GLMB parameterisations with multi-Bernoulli
births.  In the MBM filter (either labelled or not), the prediction
step is straightforward, see Section \ref{subsec:Prediction-conjugate-prior}.
This is in stark contrast with the $\delta$-GLMB filter prediction
implementation in \cite{Vo14}, which truncates the predicted density
by a $K$-shortest path algorithm. This approximation is introduced
due to an inefficient representation of the MBM. For instance, for
probability of survival lower than one, Bernoulli components that
have existence probability 1 have a smaller existence probability
after the prediction step, see Section \ref{subsec:Prediction-conjugate-prior}.
Because of this, a multi-Bernoulli density that contains $n$ Bernoulli
components, all with existence probability 1, is represented after
the prediction step by an MBM$_{01}$/$\delta$-GLMB with $2^{n}$
global hypotheses, see Proposition \ref{prop:MBM01_number_components}.
These  MBM$_{01}$/$\delta$-GLMB representations are highly inefficient
as the predicted density is simply one multi-Bernoulli process with
existence probabilities in (0,1). 

In the update step, as can be seen in Equation (\ref{eq:update_conjugate_prior2}),
we need to solve a data-association problem for each mixture component,
that is, for every global hypothesis in the prior. In this case, the
MBM parameterisation is also advantageous due to the lower number
of mixture components, compared to the MBM$_{01}$/$\delta$-GLMB
parameterisations. The reason for these advantages in the prediction
and update steps in the MBM filter is mainly due to the inefficient
MBM$_{01}$/$\delta$-GLMB parameterisations. One MBM global hypothesis
can efficiently represent many $\delta$-GLMB global hypotheses and
this extra degree of flexibility in the MBM filter simplifies the
prediction and update steps and it is independent of whether or not
we use labels. 

In addition, if there are Poisson births, the PMBM characterises the
Poisson part by its intensity, which is an efficient way of representing
a Poisson distribution. In contrast, if we were to use a labelled
Poisson process to model target births, the $\delta$-GLMB parameterisation
would need an infinite number of global hypotheses to represent the
Poisson part, since each global hypothesis  in the $\delta$-GLMB
density has a deterministic cardinality.

\section{Implementation for linear/Gaussian dynamic and measurement models\label{sec:Linear-Gaussian-implementation}}

In this section we propose an implementation of the PMBM filter for
linear Gaussian dynamic and measurement models with Poisson births.
  We first provide an overview of the structure of the hypotheses
in Section \ref{subsec:Structure-of-the-hypotheses}. Then, we explain
the prediction and update in Sections \ref{subsec:Prediction-implementation}
and \ref{subsec:Update-implementation}, respectively. 

\subsection{Structure of the hypotheses\label{subsec:Structure-of-the-hypotheses}}

\begin{figure}
\begin{centering}
\includegraphics[clip,scale=0.65]{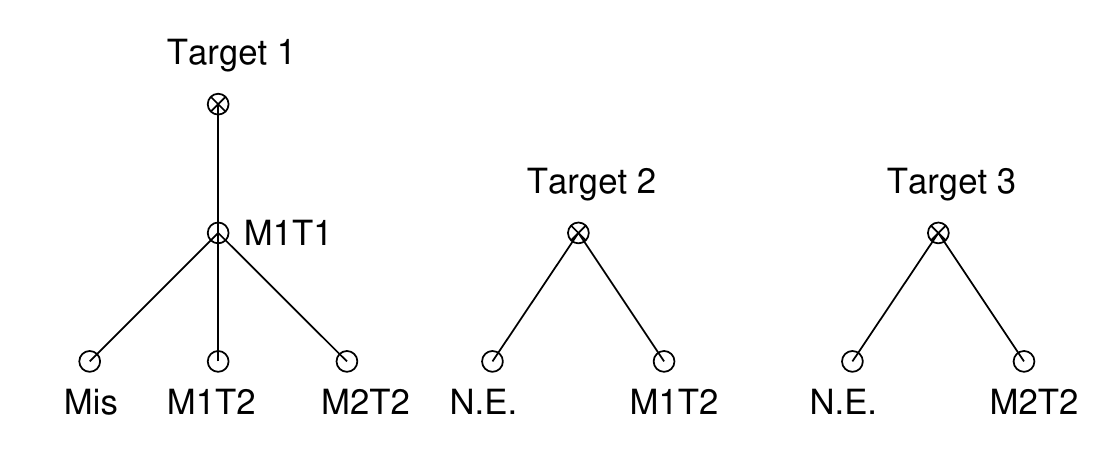}
\par\end{centering}
\caption{\label{fig:Single-target-hypothesis}Illustration of the single-target
hypothesis tree. We consider there is one measurement at time 1 (M1T1)
and two measurements at time 2 (M1T2 and M2T2). The hypothesis tree
at time 2 considers that potentially detected target 1 is associated
to M1T1 at time 1. At time 2, it can be associated with a misdetection
(Mis) or with M1T2 or M2T2. Potentially detected target 2 might not
exist (N.E.) or be associated to M1T2. Potentially detected target
3 might not exist or be associated to M2T2. There are 3 global hypotheses
at time 2. All the global hypotheses associate M1T1 to potentially
detected target 1. At time 2, the measurement associations to potentially
detected targets 1, 2 and 3 in the global hypotheses are: (Mis, M1T2,
M2T2), (M1T2, N.E, M2T2) and (M2T2,M1T2, N.E).}
\end{figure}

In the conjugate prior, see (\ref{eq:prior}), there is an index $j$
for the multi-Bernoulli mixture. Each $j$ corresponds to a global
hypothesis, which represents possible association of measurements
to potentially detected targets. As explained in \cite{Williams15b},
global hypotheses can be expressed in terms of single-target hypothesis.
A single-target hypothesis corresponds to a sequence of measurements
associated to a potentially detected target. Given a single-target
hypothesis, this potentially detected target follows a Bernoulli distribution,
as explained in Section \ref{sec:Conjugate-priors-proof}. Therefore,
each measurement starts a new single-target hypothesis. At following
time steps, new single-target hypotheses are created by associating
previous single-target hypotheses with current measurements or with
a misdetection. By doing this, global hypotheses are a collection
of these single-target hypotheses, with the conditions that no measurement
is left without being associated and a measurement can only be assigned
to one single target hypothesis. This hypothesis structure resembles
the one in track-oriented MHT \cite{Kurien_inbook90} and is illustrated
in Figure \ref{fig:Single-target-hypothesis}. We proceed to explain
the prediction and update steps.

\subsection{Prediction\label{subsec:Prediction-implementation}}

We assume that, in the posterior at the previous time step, the Poisson
component is a Gaussian mixture
\begin{align*}
\lambda^{u}\left(x\right) & =\sum_{i=1}^{N_{u}}w_{u,i}\mathcal{N}\left(x;\overline{x}_{u,i}^{p},P_{u,i}^{p}\right)
\end{align*}
and the multi-Bernoulli mixture parameters are $w_{j,i}^{u}$, $p_{j,i}^{u}\left(x\right)=\mathcal{N}\left(x;\overline{x}_{j,i}^{u},P_{j,i}^{u}\right)$,
$r_{j,i}^{u}$. 

We also assume constant probability of survival $p_{s}$, linear/Gaussian
dynamics $g\left(x\left|y\right.\right)=\mathcal{N}\left(x;Fy,Q\right)$
and new born target intensity
\begin{align*}
\lambda^{b}\left(x\right) & =\sum_{i=1}^{N_{b}}w_{b,i}^{p}\mathcal{N}\left(x;\overline{x}_{b,i}^{p},P_{b,i}^{p}\right).
\end{align*}
Then, from Section \ref{subsec:Prediction-conjugate-prior} and using
known results from the Kalman filter prediction step \cite{Sarkka_book13},
we find that the predicted intensity is a Gaussian mixture
\begin{align}
\mu\left(x\right) & =\lambda^{b}\left(x\right)+p_{s}\sum_{i=1}^{N_{u}}w_{u,i}\mathcal{N}\left(x;F\overline{x}_{u,i}^{p},FP_{u,i}^{p}F^{T}+Q\right).\label{eq:GM_Poisson_predicted}
\end{align}
The predicted Bernoulli components have the same weights as in the
previous time step with existence $r_{j,i}=r_{j,i}^{u}p_{s}$ and
\begin{align*}
p_{j,i}\left(x\right) & =\mathcal{N}\left(x;F\overline{x}_{j,i}^{u},FP_{j,i}^{u}F^{T}+Q\right).
\end{align*}

Clearly, the implementation of the prediction step is straightforward,
contrary to the prediction step of the $\delta$-GLMB filter in \cite{Vo14},
as discussed in Section \ref{subsec:Discussion}.

\subsection{Update\label{subsec:Update-implementation}}

We assume that $p_{d}$ is constant and $p\left(z|x\right)=\mathcal{N}\left(z;Hx,R\right)$.
We rewrite the predicted intensity of the Poisson part (\ref{eq:GM_Poisson_predicted})
as 
\begin{align}
\mu\left(x\right) & =\sum_{i=1}^{N_{\mu}}w_{\mu,i}\mathcal{N}\left(x;\overline{x}_{\mu,i},P_{\mu,i}\right)\label{eq:GM_Poisson_prior}
\end{align}
and the multi-Bernoulli mixture parameters as $w_{j,i}$, $p_{j,i}\left(x\right)=\mathcal{N}\left(x;\overline{x}_{j,i},P_{j,i}\right)$,
$r_{j,i}$. 

From the conjugate prior update, see Section \ref{subsec:Update-conjugate-prior2},
we have that three different types of updates: update for undetected
targets (Poisson component), update for potential targets detected
for the first time and update for previously potentially detected
targets. The update of the Poisson component is straightforward. Using
(\ref{eq:undetected_targets}), the updated intensity for undetected
targets is (\ref{eq:GM_Poisson_prior}) multiplied by $1-p_{d}$.
We proceed to explain the other two updates.

\subsubsection{Potential targets detected for the first time\label{subsec:Targets-detected-first-time-implementation}}

We first go through all components of the Poisson prior and perform
ellipsoidal gating \cite{Kurien_inbook90} on the measurements to
lower the computational complexity. For those measurements that can
create a new track according to the gating output, we perform the
Bayesian update (\ref{eq:Bernoulli_Poisson_component1}). For measurement
$z$, this gives a Bernoulli component with existence $r^{p}\left(z\right)$
and target state density $p^{p}\left(x|z\right)$ such that
\begin{align}
r^{p}\left(z\right) & =e\left(z\right)/\rho^{p}\left(z\right)\label{eq:target_first_time_existence}\\
p^{p}\left(y|z\right) & =p_{d}p\left(z|y\right)\mu\left(y\right)/e\left(z\right)\nonumber \\
 & =\sum_{i=1}^{N_{\mu}}w_{i}\left(z\right)\mathcal{N}\left(x;\overline{x}_{\mu,i}^{u}\left(z\right),P_{\mu,i}^{u}\right)\label{eq:target_first_time_pdf_mixture}
\end{align}
where
\begin{align}
e\left(z\right) & =p_{d}\int p\left(z|y\right)\mu\left(y\right)dy\nonumber \\
 & =p_{d}\sum_{i=1}^{N_{\mu}}w_{\mu,i}\mathcal{N}\left(z;H\overline{x}_{\mu,i},S_{\mu,i}\right)\nonumber \\
\rho^{p}\left(z\right) & =e\left(z\right)+c\left(z\right)\label{eq:ro_target_first_time_existence}\\
w_{i}\left(z\right) & \propto w_{\mu,i}\mathcal{N}\left(z;H\overline{x}_{\mu,i},S_{\mu,i}\right)\nonumber \\
\overline{x}_{\mu,i}^{u}\left(z\right) & =\overline{x}_{\mu,i}+\Psi_{\mu,i}S_{\mu,i}^{-1}\left(z-H\overline{x}_{\mu,i}\right)\nonumber \\
P_{\mu,i}^{u} & =P_{\mu,i}-\Psi_{\mu,i}S_{\mu,i}^{-1}\Psi_{\mu,i}^{T}\nonumber \\
\Psi_{\mu,i} & =P_{\mu,i}H^{T}\nonumber \\
S_{\mu,i} & =HP_{\mu,i}H^{T}+R\nonumber 
\end{align}
and we recall that $c\left(\cdot\right)$ is the clutter intensity.
Note that $\overline{x}_{\mu,i}^{u}\left(z\right),P_{\mu,i}^{u}$
are the updated mean and covariance matrix of a Kalman filter with
prior $\overline{x}_{\mu,i}$ and $P_{\mu,i}$ \cite{Sarkka_book13}.
For computational complexity, we approximate the Gaussian mixture
in (\ref{eq:target_first_time_pdf_mixture}) as a Gaussian by performing
moment matching. 

We still have to determine the hypothesis weight of the newly created
components of the multi-Bernoulli mixture. According to (\ref{eq:update_conjugate_prior2}),
the hypothesis weight $w_{j,i}$ of a potential target detected for
the first time with measurement $z$ in a global hypothesis $j$ that
considers it is $\rho^{p}\left(z\right)$, which is given by (\ref{eq:ro_target_first_time_existence}).
If the global hypothesis $j$ does not consider this potentially detected
target $w_{j,i}=1$ and its existence probability is set to zero. 

\subsubsection{Previous potentially detected targets\label{subsec:Previously-detected-implementation}}

 According to Section \ref{subsec:Update-of-one-Bernoulli}, we go
through all potentially detected targets and their single target hypotheses
in (\ref{eq:multiBernoullimixture_multiplicative}) and create the
new single target hypotheses. In order to explain this procedure,
let us consider that a single target hypothesis with indices $j,i$
which has weight $w_{j,i}$, existence probability $r_{j,i}$ and
Gaussian density for the target
\begin{align}
p_{j,i}\left(x\right) & =\mathcal{N}\left(x;\overline{x}_{j,i},P_{j,i}\right).\label{eq:implementation_update_previous_targets}
\end{align}
For this single target hypothesis, we first create a new misdetection
hypothesis, which has a weight $w_{j,i}\left(1-r_{j,i}+r_{j,i}\left(1-p_{d}\right)\right)$.
The associated Bernoulli component has an existence probability $r_{j,i}\left(1-p_{d}\right)/\left(1-r_{j,i}+r_{j,i}\left(1-p_{d}\right)\right)$
and the density given that the target exists remains the same, $p_{j,i}\left(\cdot\right)$.
We then perform ellipsoidal gating \cite{Kurien_inbook90} using (\ref{eq:implementation_update_previous_targets})
to consider only the relevant measurements. For each of the chosen
measurements and this Bernoulli component, we perform the update (\ref{eq:update_Bernoulli_component}),
which has a closed-form expression given by the update step of the
Kalman filter\cite{Sarkka_book13}. For measurement $z$, we have
that the corresponding hypothesis weight is
\begin{align*}
 & w_{j,i}r_{j,i}p_{d}\mathcal{N}\left(z;H\overline{x}_{j,i},S_{j,i}\right)
\end{align*}
and the Bernoulli component has existence probability one and density
\begin{align*}
 & \mathcal{N}\left(x;\overline{x}_{j,i}^{u}\left(z\right),P_{j,i}^{u}\right)
\end{align*}
where
\begin{align*}
\overline{x}_{j,i}^{u}\left(z\right) & =\overline{x}_{j,i}+\Psi_{j,i}S_{j,i}^{-1}\left(z-H\overline{x}_{j,i}\right)\\
P_{j,i}^{u} & =P_{j,i}-\Psi_{j,i}S_{j,i}^{-1}\Psi_{j,i}^{T}\\
\Psi_{j,i} & =P_{j,i}H^{T}\\
S_{j,i} & =HP_{j,i}H^{T}+R.
\end{align*}

\subsubsection{Selection of $k$-best global hypotheses\label{subsec:Selection-of-k-best-implementation}}

At this point, we have calculated all possible new single-target hypotheses
but we still have to form the global hypotheses. We can see in (\ref{eq:update_conjugate_prior2})
that, for each global hypothesis $j$ at the previous time step, we
must go through all possible data association hypotheses that give
rise to the updated global hypotheses. This high increase in the number
the global hypotheses is the bottleneck of the computation of the
conjugate prior. However, based on the literature on labelled RFSs
and MHT, we approximate this update by pruning the number of hypotheses
using Murty's algorithm \cite{Murty68}. With this algorithm, we can
select the $k$ new global hypotheses with highest weight for a given
global hypothesis $j$ without evaluating all the newly generated
global hypotheses \cite{Vo13,Vo14,Cox95,Cox96}. An interesting alternative
would be to use the generalised Murty's algorithm for multiple frames
\cite{Fortunato07}.

For global hypothesis $j$, all measurements (excluding those removed
by gating) must be associated either to an existing track in hypothesis
$j$ or to a new track, i.e., no measurement is left unassigned. We
can then construct the corresponding cost matrix using the updated
weights of the conjugate prior. Let us assume there are $n_{o}$ old
tracks in global hypothesis $j$ and $m$ measurements $z_{1},...,z_{m}$
after gating. The cost matrix is
\begin{align}
C & =-\left[\begin{array}{cc}
\ln\left(W_{ot}\right), & \ln\left(W_{nt}\right)\end{array}\right]\label{eq:cost_matrix}
\end{align}
where
\begin{align*}
W_{nt} & =\mathrm{diag}\left(\rho^{p}\left(z_{1}\right),...,\rho^{p}\left(z_{m}\right)\right)
\end{align*}
with $\rho^{p}\left(z_{i}\right)$ given by (\ref{eq:ro_target_first_time_existence}).
Matrix $W_{nt}$ represents the weight matrix for new potentially
detected targets and $W_{ot}\in\mathbb{R}^{m\times n_{j}}$ represents
the weight matrix for old targets, where $n_{j}$ are the number of
potentially detected targets at the previous time steps in global
hypothesis $j$. Component $p,i$ of $W_{ot}$ represents the weight
of the $p$th measurement associated to $i$th target, which is
\begin{align*}
 & w_{j,i}\rho_{j,i}\left(\left\{ z_{p}\right\} \right)/\rho_{j,i}\left(\emptyset\right)\\
 & \quad=\frac{w_{j,i}r_{j,i}p_{d}\mathcal{N}\left(z_{p};H\overline{x}_{j,i},S_{j,i}\right)}{w_{j,i}\left(1-r_{j,i}+r_{j,i}\left(1-p_{d}\right)\right)},
\end{align*}
according to Section \ref{subsec:Previously-detected-implementation}.
Note that we normalise the previous weights by $\rho_{j,i}\left(\emptyset\right)$
so that the weight of a hypothesis that does not assign a measurement
to a target is the same for an old and a new target. This is just
done so that we can obtain the $k$-best global hypotheses efficiently
using Murty's algorithm but we do not alter the real weights, which
are unnormalised. Each new global hypothesis that originates from
hypothesis $j$ can be written as an $m\times\left(m+n_{0}\right)$
assignment matrix $S$ consisting of 0 or 1 entries such that each
row sums to one and each column sums to zero or one. Then, we select
the $k$ best global hypotheses that minimise $\mathrm{tr}\left(S^{T}C\right)$
using Murty's algorithm \cite{Murty68}. For global hypothesis $j$,
whose weight is $w_{j}\propto\prod_{i=1}^{n}w_{j,i}$, we suggest
choosing $k=\left\lceil N_{h}\cdot w_{j}\right\rceil $, where it
is assumed that we want a maximum number $N_{h}$ of global hypotheses
as in \cite{Vo14}. This way, global hypotheses with higher weights
will give rise to more global hypotheses. Note that this part of the
algorithm is quite similar to the $\delta$-GLMB filter update with
just some modifications in the cost matrix \cite[Sec. IV]{Vo14}.
Finally, the pseudo-code of a prediction and an update is given in
Algorithm \ref{alg:PMBM_pseudocode}. 

\selectlanguage{british}%
\begin{algorithm}
\selectlanguage{english}%
\caption{\foreignlanguage{british}{\label{alg:PMBM_pseudocode}\foreignlanguage{english}{Pseudo-code
for one prediction and update for PMBM filter}}}

{\fontsize{9}{9}\selectfont

\textbf{Input:} Parameters of the PMBM posterior at the previous time
step, see Section \ref{subsec:Prediction-implementation}, and measurement
set $Z$ at current time step.

\textbf{Output: }Parameters of the PMBM posterior at the current time
step. 

\begin{algorithmic}     

\State - Perform prediction, see Section \ref{subsec:Prediction-implementation}.

\State $\phantom{}$ \foreignlanguage{british}{\Comment{Update}}

\selectlanguage{british}%
\For{$z\in Z$} \Comment{Targets detected for first time}

\selectlanguage{english}%
\State - Perform ellipsoidal gating of $z$ w.r.t. Gaussian components
of Poisson prior (\ref{eq:GM_Poisson_prior}).

\If{$z$ meets ellipsoidal gating for at least one component}

\State - Create a new Bernoulli component, see Section \ref{subsec:Targets-detected-first-time-implementation}.

\EndIf

\selectlanguage{british}%
\EndFor

\For{$i=1$ to $n$} \Comment{We go through all possible targets}

\For{$j_{i}=1$ to $l_{i}$} \Comment{$l_{i}$ is the number of single-target hypotheses for possible target $i$}

\selectlanguage{english}%
\State - Create new misdetection hypothesis, see Section \ref{subsec:Previously-detected-implementation}.

\State - Perform gating on $Z$ and create new detection hypotheses,
see Section \ref{subsec:Previously-detected-implementation}.

\selectlanguage{british}%
\EndFor

\EndFor

\For{all $j$} \Comment{We go through all previous global hypotheses}

\selectlanguage{english}%
\State - Create cost matrix (\ref{eq:cost_matrix}).

\State - Run Murty's algorithm to select $k=\left\lceil N_{h}\cdot w_{j}\right\rceil $
new global hypotheses, see Section \ref{subsec:Selection-of-k-best-implementation}. 

\selectlanguage{british}%
\EndFor

\selectlanguage{english}%
\State - Estimate target states, see Section \ref{sec:Estimation}.

\State $\phantom{}$ \foreignlanguage{british}{\Comment{Pruning}}

\State - Prune the Poisson part by discarding components whose weight
is below a threshold. 

\State - Prune global hypotheses by keeping the highest $N_{h}$
global hypotheses.

\State - Remove Bernoulli components whose existence probability
is below a threshold or do not appear in the pruned global hypotheses. 

\selectlanguage{british}%
\end{algorithmic}

}
\end{algorithm}

\selectlanguage{english}%

\section{Estimation\label{sec:Estimation}}

In this section, we discuss how to perform target state estimation
in the PMBM filter. In a multiple target system, an optimal estimator
is given by minimising a multi-target metric, for example, the optimal
subpattern assignment (OSPA) metric \cite{Schuhmacher08,Guerriero10,Williams15}.
Nevertheless, there are suboptimal estimators that are easy to compute
and can work very well in many cases. In this section, we provide
tractable methods for obtaining the (suboptimal) estimators used in
MHT (Estimator 3) and the $\delta$-GLMB filter (Estimator 2) using
the PMBM distribution form. We also propose an additional estimator
based on the PMBM (Estimator 1).

\subsection{Estimator 1}

In Estimator 1, we first select the global hypothesis of the multi-Bernoulli
mixture in (\ref{eq:multiBernoullimixture_multiplicative}) with highest
weight, which corresponds to obtaining index
\[
j^{*}=\arg\max_{j}\prod_{i=1}^{n}w_{j,i}.
\]
Then, we report the mean of the Bernoulli components in hypothesis
$j^{*}$ whose existence probability is above a threshold $\Gamma$.
Given the probabilities of detection and survival, this threshold
determines the number of consecutive misdetections we can have from
a target to report its estimate, see prediction and update for missed
targets in Sections \ref{subsec:Prediction-conjugate-prior} and \ref{subsec:Update-of-one-Bernoulli}.

\subsection{Estimator 2}

Estimator 2 is the same kind of estimator as the one proposed in the
$\delta$-GLMB filter \cite{Vo14}, which we proceed to describe.
The $\delta$-GLMB filter estimator first obtains the maximum a posteriori
(MAP) estimate of the cardinality. Then, it finds the global hypothesis
with this cardinality with highest weight and reports the mean of
the targets in this hypothesis. 

The same type of estimate can be constructed from the multi-Bernoulli
mixture in (\ref{eq:prior_prev}) by first calculating its cardinality
distribution \cite[Eq. (11.115)]{Mahler_book07}
\begin{equation}
p(n)\propto\sum_{j}\left[\prod_{i}w_{j,i}\right]p_{j}(n)\label{eq:MBMCardinality}
\end{equation}
where $p_{j}(n)$ is the cardinality distribution of term $j$ of
the mixture. The cardinality distribution $p_{j}(n)$ can be calculated
efficiently using a discrete Fourier transform as the cardinality
distribution of a multi-Bernoulli RFS is the convolution of the cardinality
distributions of its Bernoulli components \cite{Fernandez10}. By
finding the value of $n$ that maximises (\ref{eq:MBMCardinality}),
we obtain the MAP cardinality $n^{*}$. We can then obtain the highest
weight global hypothesis with deterministic cardinality, implicitly
represented by the multi-Bernoulli mixture, from the global hypothesis
\begin{equation}
j^{*}=\arg\max_{j}\prod_{l=1}^{n^{*}}w_{j,i_{l}}r_{j,i_{l}}\prod_{l=n^{*}+1}^{n}w_{j,i_{l}}(1-r_{j,i_{l}})\label{eq:delta_GLMB_best_hypothesis}
\end{equation}
where $i_{1},\dots,i_{n}$ is an ordering such that $r_{j,i_{l}}\geq r_{j,i_{l+1}}\;\forall\;l$.
 Note that given a MBM hypothesis $j$, the weight of the deterministic
hypothesis with highest weight is given by the term inside the argmax
in (\ref{eq:delta_GLMB_best_hypothesis}), Once we have found the
global hypothesis $j^{*}$, the set estimate is formed by the means
of the $n^{*}$ Bernoulli components with highest existence in this
hypothesis. 

\subsection{Estimator 3}

Estimator 3 is the same type of estimator as the one proposed in the
MHT of \cite{Reid79,Mori86}, which has also been suggested for the
$\delta$-GLMB filter \cite{Vo14}. This estimate first obtains the
global hypothesis with a deterministic cardinality with highest weight,
i.e., the MAP estimate of the global hypotheses with deterministic
cardinality. Note that the global hypotheses (and their weights) with
deterministic cardinality (no uncertainty in the cardinality distribution)
can be obtained from the multi-Bernoulli mixture (\ref{eq:multiBernoullimixture_multiplicative})
by expanding each Bernoulli component so that, in each of the resulting
mixture components, either a target exists or not. Then, the estimate
is constructed by reporting the mean of the targets in this hypothesis. 

We proceed to explain how to obtain this kind of estimate directly
from the multi-Bernoulli mixture. We obtain the MAP estimate of the
global hypotheses with deterministic cardinality by finding
\begin{equation}
j^{*}=\arg\max_{j}\prod_{i|r_{j,i}\geq0.5}w_{j,i}r_{j,i}\prod_{i|r_{j,i}<0.5}w_{j,i}(1-r_{j,i}).\label{eq:MHT_best_hypothesis}
\end{equation}
It should be noted that the term inside the argmax in (\ref{eq:MHT_best_hypothesis})
corresponds to the the weight of the deterministic hypothesis with
highest weight for the $j$th MBM hypothesis. The set estimate is
formed by the means of the Bernoulli components for global hypothesis
$j^{*}$ whose existences are above 0.5, as indicated in (\ref{eq:MHT_best_hypothesis}).
In summary, we find that both the $\delta$-GLMB style and the MHT
style estimators can be easily constructed from the multi-Bernoulli
mixture representation. 

\section{Simulations\label{sec:Simulations}}

In this section, we show simulation results that compare the PMBM
filter with the Gaussian mixture PHD, CPHD filters \cite{Vo06,Vo07}
and, track-oriented and measurement-oriented multi-Bernoulli/Poisson
(TOMB/MOMB) filters in \cite{Williams15b}. We also analyse the behaviours
of the three estimators proposed in Section \ref{sec:Estimation}.
We consider an area $\left[0,300\right]\times\left[0,300\right]$
and all the units in this section are in international system. Target
states consist of 2D position and velocity $\left[p_{x},v_{x},p_{y},v_{y}\right]^{T}$
and are born according to a Poisson process of intensity 0.005 and
Gaussian density with mean $\left[100,0,100,0\right]^{T}$ and covariance
$\mathrm{diag}\left(\left[150^{2},1,150^{2},1\right]\right)$, which
covers the region of interest. We use the following parameters for
the simulation:
\begin{align*}
F=I_{2}\otimes\left(\begin{array}{cc}
1 & T\\
0 & 1
\end{array}\right), & \,Q=qI_{2}\otimes\left(\begin{array}{cc}
T^{3}/3 & T^{2}/2\\
T^{2}/2 & T
\end{array}\right)\\
H=I_{2}\otimes\left(\begin{array}{cc}
1 & 0\end{array}\right), & \,R=I_{2}
\end{align*}
where $\otimes$ is the Kronecker product, $q=0.01$, $T=1$, $p_{s}=0.99$.
We also consider Poisson clutter uniform in the region of interest
with $\lambda_{c}=10$, which implies 10 expected false alarms per
time step, and $p_{d}=0.9$. The filters consider that there are no
targets at time 0.

The PMBM filter implementation uses a maximum number of global hypotheses
$N_{h}=200$, estimation threshold for estimator 1 is $\Gamma=0.4$,
which allows two consecutive misdetections for $p_{d}=0.9$ and $p_{s}=0.99$
to report an estimate, see Section \ref{sec:Estimation}. In the Poisson
part, we use a pruning threshold of $10^{-5}$. For the MB part, we
remove Bernoulli components whose existence probability is lower than
$10^{-5}$. We also use ellipsoidal gating \cite{Kurien_inbook90}
with threshold 20.  TOMB/MOMB report estimates for targets with existence
probability higher than 0.7. 

\begin{figure}
\begin{centering}
\includegraphics[scale=0.6]{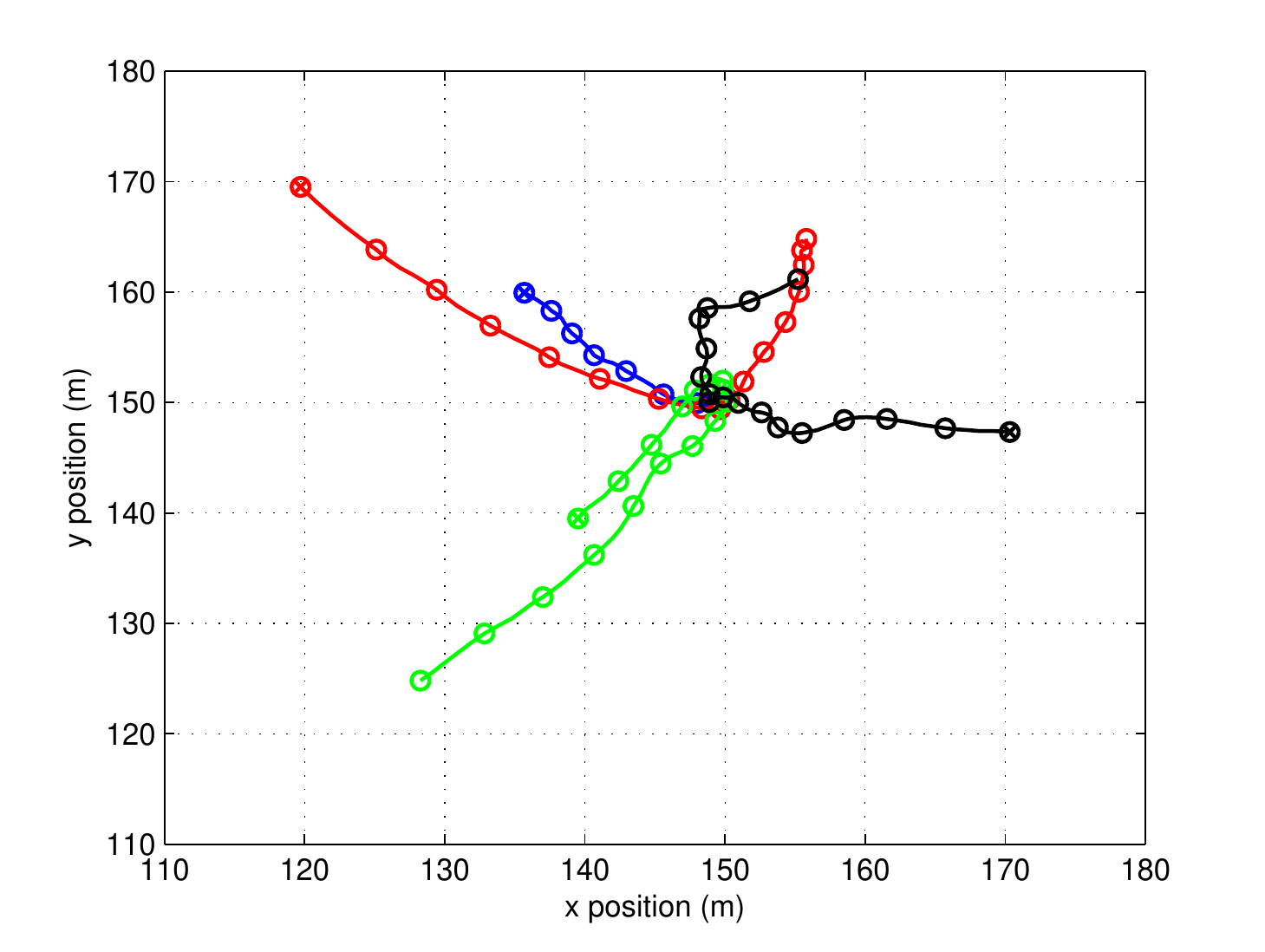}
\par\end{centering}
\caption{\label{fig:Scenario}Scenario of simulations. There are four targets,
all born at time step 1 and alive throughout the simulation, except
the blue target that dies at time step 40, when all targets are in
close proximity. Initial target positions have a cross and target
positions every 5 time steps have a circle. }
\end{figure}

\begin{figure}
\begin{centering}
\includegraphics[scale=0.6]{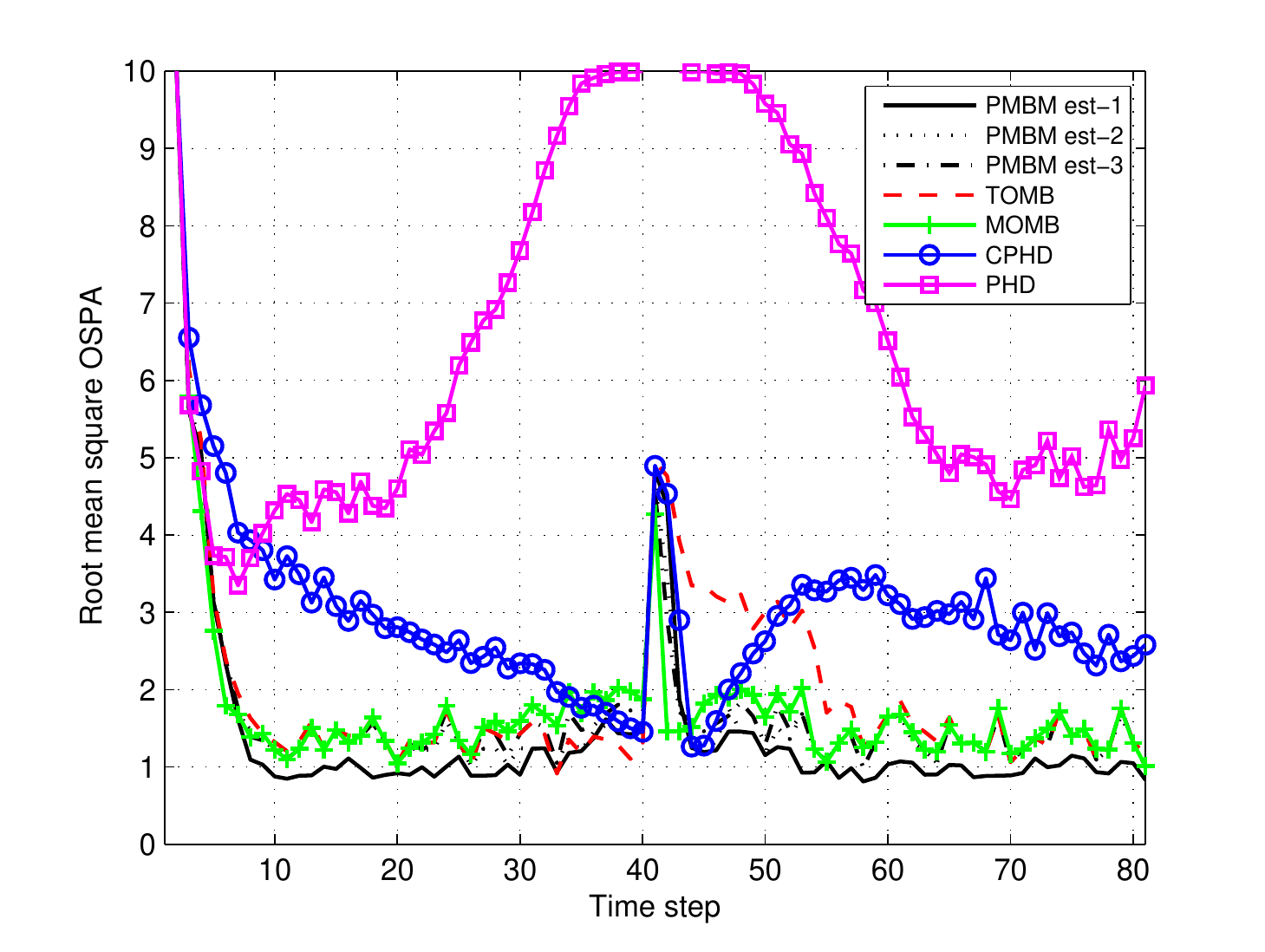}
\par\end{centering}
\caption{\label{fig:Mean-OSPA-error}Mean OSPA error for the algorithms for
$p_{d}=0.9$ and . The PMBM filter outperforms the rest of the algorithms.
Estimator 1 of the PMBM filter provides lowest error and Estimators
2 and 3 perform similarly. }

\end{figure}

We consider 81 time steps and the scenario in Figure \ref{fig:Scenario}.
These trajectories were generated as indicated in \cite[Sec. VI]{Williams15b}.
For each trajectory, we initiate the midpoint (state at time step
41) from a Gaussian with mean $\left[150,0,150,0\right]^{T}$ and
covariance matrix $0.1I_{4}$ and the rest of the trajectory is generated
running forward and backward dynamics. This scenario is challenging
due to the broad Poisson prior that covers the region of interest,
the high number of targets in close proximity and the fact that one
target dies when they are in close proximity. We perform 100 Monte
Carlo runs and obtain the root mean square optimal subpattern assignment
(OSPA) error $\left(p=2,\,c=10\right)$ \cite{Schuhmacher08,Rahmathullah17}
at each time step for each algorithm, as shown in Figure \ref{fig:Mean-OSPA-error}.
Estimator 1 applied to the PMBM filter provides the lowest errors
followed by Estimators 2 and 3, which behave similarly. MOMB performs
as accurately as Estimators 2 and 3 of the PMBM. It takes TOMB a long
time to determine that one target disappears at time step 40. PHD
and CPHD are rougher approximations and do not perform well in this
scenario. 

We also show the root mean square OSPA error averaged over all time
steps of the algorithms for different values of $p_{d}$ and $\lambda_{c}=10$
in Table \ref{tab:Root-mean-square}. On the whole, the PMBM filter
performs better than the rest regardless of the estimator. Estimator
1 has lower error than Estimator 2 and 3 for $p_{d}$ equal or higher
than 0.9. For lower values of $p_{d}$, Estimator 2 provides lowest
errors. The MOMB has the second best performance followed by the TOMB
algorithm. The CPHD and PHD filters perform much worse than the other
filters. 

\begin{table*}
\caption{\label{tab:Root-mean-square}Root mean square OSPA error for the algorithms
at all time steps}
\begin{centering}
\begin{tabular}{c|ccccccc}
\hline 
$\left(p_{d},\lambda_{c}\right)$ &
PMBM Est 1 &
PMBM Est 2 &
PMBM Est 3 &
TOMB &
MOMB &
CPHD &
PHD\tabularnewline
\hline 
$\left(0.95,10\right)$ &
\textbf{2.10} &
\textbf{2.10} &
\textbf{2.10} &
2.32 &
\textbf{2.10} &
2.83 &
6.34\tabularnewline
$\left(0.95,15\right)$ &
\textbf{2.15} &
2.17 &
\textbf{2.15} &
2.48 &
2.17 &
2.97 &
6.44\tabularnewline
$\left(0.95,20\right)$ &
\textbf{2.26} &
2.27 &
\textbf{2.26} &
2.61 &
2.27 &
3.00 &
6.51\tabularnewline
$\left(0.9,10\right)$ &
\textbf{2.23} &
2.34 &
2.36 &
2.65 &
2.37 &
3.39 &
7.05\tabularnewline
$\left(0.9,15\right)$ &
\textbf{2.30} &
2.42 &
2.44 &
2.75 &
2.45 &
3.45 &
7.04\tabularnewline
$\left(0.9,20\right)$ &
\textbf{2.37} &
2.48 &
2.50 &
2.80 &
2.53 &
2.57 &
7.18\tabularnewline
$\left(0.8,10\right)$ &
2.67 &
\textbf{2.64} &
2.66 &
2.95 &
2.78 &
4.19 &
8.22\tabularnewline
$\left(0.8,15\right)$ &
2.80 &
\textbf{2.78} &
2.80 &
3.15 &
2.88 &
4.25 &
8.23\tabularnewline
$\left(0.8,20\right)$ &
2.93 &
\textbf{2.90} &
2.92 &
3.18 &
3.00 &
4.48 &
8.34\tabularnewline
$\left(0.7,10\right)$ &
3.02 &
\textbf{2.99} &
3.01 &
3.47 &
3.15 &
4.83 &
8.80\tabularnewline
$\left(0.7,15\right)$ &
3.10 &
\textbf{3.07} &
3.09 &
3.57 &
3.24 &
4.99 &
8.86\tabularnewline
$\left(0.7,20\right)$ &
3.29 &
\textbf{3.25} &
3.28 &
3.67 &
3.41 &
5.09 &
8.87\tabularnewline
$\left(0.6,10\right)$ &
3.42 &
\textbf{3.39} &
3.42 &
3.81 &
3.55 &
5.30 &
9.09\tabularnewline
$\left(0.6,15\right)$ &
3.62 &
\textbf{3.60} &
3.62 &
4.03 &
3.72 &
5.52 &
9.14\tabularnewline
$\left(0.6,20\right)$ &
3.71 &
\textbf{3.69} &
3.71 &
4.09 &
3.82 &
5.61 &
9.18\tabularnewline
\hline 
\end{tabular}
\par\end{centering}
\begin{centering}
\par\end{centering}
\end{table*}

\section{Conclusions\label{sec:Conclusions}}

In this paper, we have first provided a non-PGFL derivation of the
Poisson multi-Bernoulli mixture filter in \cite{Williams15b}, showing
its conjugacy property. In order to attain this, we have used a suitable
representation of the prior density, which is the union of a Poisson
and a multi-Bernoulli mixture, as well as different representations
of the likelihood function at several steps. In addition, we have
also proved that this derivation can be directly extended to the labelled
case by removing the Poisson component and adding unique labels to
the Bernoulli components. We have also explained that the PMBM filter
parameterisation has important benefits compared to the $\delta$-GLMB
filter parameterisation, which considers hypotheses with deterministic
cardinality.

We have also provided an implementation of the Poisson multi-Bernoulli
mixture filter for linear/Gaussian measurement models and Poisson
births and clutter. The multi-Bernoulli mixture is a more efficient
parameterisation of the filtering density than the $\delta$-GLMB
form and, consequently, the prediction step is greatly simplified.
Based on the multiple target tracking literature on MHT and labelled
random finite sets, we have suggested three suboptimal estimators
for the PMBM filter and how they can be obtained efficiently. Finally,
we have compared the performance of the PMBM filter with other RFS
filters in a challenging scenario, in which new born targets are distributed
according to a Poisson RFS with an intensity that covers the surveillance
area and several targets get in close proximity. PMBM outperforms
the rest of the filters in this scenario.

\appendices{}

\section{\label{sec:AppendixA}}

In this appendix, we prove (\ref{eq:likelihood_new_form}). We denote
\begin{align}
l_{s}\left(\left\{ z_{1},...,z_{m}\right\} |X\right) & =e^{-\lambda_{c}}\sum_{U\uplus Y_{1}...\uplus Y_{m}=X}\left[1-p_{d}\left(\cdot\right)\right]^{U}\nonumber \\
 & \quad\times\prod_{i=1}^{m}\tilde{l}\left(z_{i}|Y_{i}\right).\label{eq:likelihood_new_form_appendix}
\end{align}
We perform a proof by induction. In the rest of this appendix, we
denote $Z=\left\{ z_{1},...,z_{m}\right\} $ and $X=\left\{ x_{1},...,x_{n}\right\} $
for notational simplicity. First, we note that
\begin{align}
l\left(\emptyset|\emptyset\right) & =l_{s}\left(\emptyset|\emptyset\right)=e^{-\lambda_{c}}.\label{eq:initial_proof}
\end{align}
The result is proved if we prove that
\begin{align}
l\left(\left\{ z_{1},...,z_{j}\right\} |\left\{ x_{1},...,x_{i}\right\} \right) & =l_{s}\left(\left\{ z_{1},...,z_{j}\right\} |\left\{ x_{1},...,x_{i}\right\} \right)\label{eq:Equivalence_likelihood_proof}
\end{align}
for $j\leq m$ and $i\leq n$, implies that
\begin{align}
l\left(Z\uplus\left\{ z_{m+1}\right\} |X\right) & =l_{s}\left(Z\uplus\left\{ z_{m+1}\right\} |X\right)\label{eq:likelihood_proof_result1}
\end{align}
and
\begin{align}
l\left(Z|X\uplus\left\{ x_{n+1}\right\} \right) & =l_{s}\left(Z|X\uplus\left\{ x_{n+1}\right\} \right).\label{eq:likelihood_proof_result2}
\end{align}

\subsection{First part}

We proceed to prove (\ref{eq:likelihood_proof_result1}). We have
that
\begin{align}
 & l_{s}\left(Z\uplus\left\{ z_{m+1}\right\} |X\right)\nonumber \\
 & =e^{-\lambda_{c}}\sum_{U\uplus Y_{1}...\uplus Y_{m}\uplus Y_{m+1}=X}\left[1-p_{d}\left(\cdot\right)\right]^{U}\prod_{i=1}^{m+1}\tilde{l}\left(z_{i}|Y_{i}\right)\nonumber \\
 & =e^{-\lambda_{c}}\sum_{Y_{m+1}\subseteq X}\tilde{l}\left(z_{m+1}|Y_{m+1}\right)\sum_{U\uplus Y_{1}\uplus...\uplus Y_{m}=X\setminus Y_{m+1}}\nonumber \\
 & \;\times\left[1-p_{d}\left(\cdot\right)\right]^{U}\prod_{i=1}^{m}\tilde{l}\left(z_{i}|Y_{i}\right)\nonumber \\
 & =\sum_{Y_{m+1}\subseteq X}\tilde{l}\left(z_{m+1}|Y_{m+1}\right)l_{s}\left(Z|X\setminus Y_{m+1}\right)\nonumber \\
 & =\tilde{l}\left(z_{m+1}|\emptyset\right)l_{s}\left(Z|X\right)+\sum_{j=1}^{n}\tilde{l}\left(z_{m+1}|\left\{ x_{j}\right\} \right)l_{s}\left(Z|X\setminus\left\{ x_{j}\right\} \right).\label{eq:proof1_1}
\end{align}
We also have
\begin{align}
 & l\left(Z\uplus\left\{ z_{m+1}\right\} |X\right)\nonumber \\
 & =e^{-\lambda_{c}}\sum_{Z^{c}\uplus Z_{1}...\uplus Z_{n}=Z\uplus\left\{ z_{m+1}\right\} }\left[c\left(\cdot\right)\right]^{Z^{c}}\prod_{i=1}^{n}\hat{l}\left(Z_{i}|x_{i}\right)\nonumber \\
 & =e^{-\lambda_{c}}\left[\sum_{Z^{c}\uplus Z_{1}...\uplus Z_{n}=Z\uplus\left\{ z_{m+1}\right\} :z_{m+1}\in Z^{c}}\left[c\left(\cdot\right)\right]^{Z^{c}}\prod_{i=1}^{n}\hat{l}\left(Z_{i}|x_{i}\right)\right.\nonumber \\
 & \;\left.+\sum_{j=1}^{n}\sum_{Z^{c}\uplus Z_{1}...\uplus Z_{n}=Z\uplus\left\{ z_{m+1}\right\} :z_{m+1}\in Z_{j}}\left[c\left(\cdot\right)\right]^{Z^{c}}\prod_{i=1}^{n}\hat{l}\left(Z_{i}|x_{i}\right)\right]\nonumber \\
 & =e^{-\lambda_{c}}\left[\tilde{l}\left(z_{m+1}|\emptyset\right)\sum_{Z^{c}\uplus Z_{1}...\uplus Z_{n}=Z}\left[c\left(\cdot\right)\right]^{Z^{c}}\prod_{i=1}^{n}\hat{l}\left(Z_{i}|x_{i}\right)\right.\nonumber \\
 & \;+\sum_{j=1}^{n}\hat{l}\left(\left\{ z_{m+1}\right\} |x_{j}\right)\sum_{Z^{c}\uplus Z_{1}...\uplus Z_{n}=Z:Z_{j}=\emptyset}\left[c\left(\cdot\right)\right]^{Z^{c}}\nonumber \\
 & \left.\;\times\prod_{i=1:i\neq j}^{n}\hat{l}\left(Z_{i}|x_{i}\right)\right]\nonumber \\
 & =\tilde{l}\left(z_{m+1}|\emptyset\right)l\left(Z|X\right)+\sum_{i=1}^{n}\tilde{l}\left(z_{m+1}|\left\{ x_{i}\right\} \right)l\left(Z|X\setminus\left\{ x_{i}\right\} \right).\label{eq:proof1_2}
\end{align}
Using the induction hypothesis (\ref{eq:Equivalence_likelihood_proof}),
(\ref{eq:proof1_2}) equals (\ref{eq:proof1_1}), so we finish the
proof of (\ref{eq:likelihood_proof_result1}). 

\subsection{Second part\label{subsec:Second-part-AppendixA}}

We proceed to prove (\ref{eq:likelihood_proof_result2}). In this
part, we denote $p_{d}'\left(\cdot\right)=1-p_{d}\left(\cdot\right)$.
We have that
\begin{align}
 & l_{s}\left(Z|X\uplus\left\{ x_{n+1}\right\} \right)\nonumber \\
 & =e^{-\lambda_{c}}\sum_{U\uplus Y_{1}...\uplus Y_{m}=X\uplus\left\{ x_{n+1}\right\} }\left[p_{d}'\left(\cdot\right)\right]^{U}\prod_{i=1}^{m}\tilde{l}\left(z_{i}|Y_{i}\right)\nonumber \\
 & =e^{-\lambda_{c}}\left[\sum_{U\uplus Y_{1}...\uplus Y_{m}=X\uplus\left\{ x_{n+1}\right\} :x_{n+1}\in U}\left[p_{d}'\left(\cdot\right)\right]^{U}\prod_{i=1}^{m}\tilde{l}\left(z_{i}|Y_{i}\right)\right.\nonumber \\
 & \;\left.+\sum_{j=1}^{m}\sum_{U\uplus Y_{1}...\uplus Y_{m}=X\uplus\left\{ x_{n+1}\right\} :x_{n+1}\in Y_{j}}\left[p_{d}'\left(\cdot\right)\right]^{U}\prod_{i=1}^{m}\tilde{l}\left(z_{i}|Y_{i}\right)\right]\nonumber \\
 & =e^{-\lambda_{c}}\left[p_{d}'\left(x_{n+1}\right)\sum_{U\uplus Y_{1}...\uplus Y_{m}=X}\left[p_{d}'\left(\cdot\right)\right]^{U}\prod_{i=1}^{m}\tilde{l}\left(z_{i}|Y_{i}\right)\right.\nonumber \\
 & \;+\sum_{j=1}^{m}\tilde{l}\left(z_{j}|\left\{ x_{n+1}\right\} \right)\sum_{U\uplus Y_{1}...\uplus Y_{m}=X:Y_{j}=\emptyset}\left[p_{d}'\left(\cdot\right)\right]^{U}\nonumber \\
 & \left.\;\times\prod_{i=1:i\neq j}^{n}\tilde{l}\left(z_{i}|Y_{i}\right)\right]\nonumber \\
 & =p_{d}'\left(x_{n+1}\right)l_{s}\left(Z|X\right)+\sum_{j=1}^{m}\tilde{l}\left(z_{j}|\left\{ x_{n+1}\right\} \right)l_{s}\left(Z\setminus\left\{ z_{j}\right\} |X\right).\label{eq:proof2_1}
\end{align}
We also have that
\begin{align}
 & l\left(Z|X\uplus\left\{ x_{n+1}\right\} \right)\nonumber \\
 & =e^{-\lambda_{c}}\sum_{Z^{c}\uplus Z_{1}...\uplus Z_{n+1}=Z}\left[c\left(\cdot\right)\right]^{Z^{c}}\prod_{i=1}^{n+1}\hat{l}\left(Z_{i}|x_{i}\right)\nonumber \\
 & =e^{-\lambda_{c}}\sum_{Z_{n+1}\subseteq Z}\hat{l}\left(Z_{n+1}|x_{n+1}\right)\nonumber \\
 & \;\times\sum_{Z^{c}\uplus Z_{1}...\uplus Z_{n}=Z\setminus Z_{n+1}}\left[c\left(\cdot\right)\right]^{Z^{c}}\prod_{i=1}^{n}\hat{l}\left(Z_{i}|x_{i}\right)\nonumber \\
 & =e^{-\lambda_{c}}\left[p_{d}'\left(x_{n+1}\right)\sum_{Z^{c}\uplus Z_{1}...\uplus Z_{n}=Z}\left[c\left(\cdot\right)\right]^{Z^{c}}\prod_{i=1}^{n}\hat{l}\left(Z_{i}|x_{i}\right)\right.\nonumber \\
 & \left.\;+\sum_{j=1}^{m}\tilde{l}\left(z_{j}|\left\{ x_{n+1}\right\} \right)\sum_{Z^{c}\uplus Z_{1}...\uplus Z_{n}=Z\setminus\left\{ z_{j}\right\} }\left[c\left(\cdot\right)\right]^{Z^{c}}\prod_{i=1}^{n}\hat{l}\left(Z_{i}|x_{i}\right)\right]\nonumber \\
 & =p_{d}'\left(x_{n+1}\right)l\left(Z|X\right)+\sum_{j=1}^{m}\tilde{l}\left(z_{j}|\left\{ x_{n+1}\right\} \right)l\left(Z\setminus\left\{ z_{j}\right\} |X\right).\label{eq:proof2_2}
\end{align}
Given that the induction hypothesis (\ref{eq:Equivalence_likelihood_proof})
holds, (\ref{eq:proof2_1}) and (\ref{eq:proof2_2}) are identical,
so we finish the proof of (\ref{eq:likelihood_proof_result2}).

\section{\label{sec:AppendixB}}

In this appendix, we show how to update a Poisson prior, whose result
is given in (\ref{eq:update_Poisson1})-(\ref{eq:Bernoulli_Poisson_last}).
Substituting (\ref{eq:likelihood_new_form}) into (\ref{eq:BayesPoisson}),
we find
\begin{align*}
 & q^{p}\left(X|Z\right)\\
 & \propto f^{p}\left(X\right)\sum_{U\uplus Y_{1}...\uplus Y_{m}=X}\left[1-p_{d}\left(\cdot\right)\right]^{U}\prod_{i=1}^{m}\tilde{l}\left(z_{i}|Y_{i}\right)\\
 & =\sum_{U\uplus Y_{1}...\uplus Y_{m}=X}\left[1-p_{d}\left(\cdot\right)\right]^{U}\left[\prod_{i=1}^{m}\tilde{l}\left(z_{i}|Y_{i}\right)\right]\\
 & \quad\times f^{p}\left(U\uplus Y_{1}...\uplus Y_{m}\right)\\
 & \propto\sum_{U\uplus Y_{1}...\uplus Y_{m}=X}\left[1-p_{d}\left(\cdot\right)\right]^{U}f^{p}\left(U\right)\left[\prod_{i=1}^{m}\tilde{l}\left(z_{i}|Y_{i}\right)f^{p}\left(Y_{i}\right)\right]\\
 & \propto\sum_{U\uplus Y_{1}\uplus...\uplus Y_{m}=X}q^{p}\left(U\right)\left[\prod_{i=1}^{m}\rho^{p}\left(z_{i}\right)q^{p}\left(Y_{i}|z_{i}\right)\right].
\end{align*}
In the previous derivation, we have used that $f^{p}\left(U\uplus Y_{1}...\uplus Y_{m}\right)\propto f^{p}\left(U\right)\prod_{i=1}^{m}f^{p}\left(Y_{i}\right)$,
see (\ref{eq:Poisson_prior}), and Equations (\ref{eq:undetected_targets})
and (\ref{eq:Bernoulli_Poisson_component1}). The specific form of
$q^{p}\left(Y_{i}|z_{i}\right)$, which is given in (\ref{Bernoulli_Poisson_component1_evaluation}),
is obtained straightforwardly by calculating (\ref{eq:Bernoulli_Poisson_component1}).

\section{\label{sec:AppendixC}}

In this appendix, we prove (\ref{eq:likelihood_equivalence_proof}).
By definition, we know that (\ref{eq:likelihood_equivalence_proof})
is met for $n=0$ as $l_{o}\left(Z|Y\right)=l\left(Z|Y\right)$. By
induction, Equation (\ref{eq:likelihood_equivalence_proof}) is proved
if the equality 
\begin{align*}
l_{o}\left(Z|Y,X_{1},...,X_{n}\right) & =l\left(Z|Y\uplus X_{1}\uplus...\uplus X_{n}\right)
\end{align*}
implies 
\begin{align*}
l_{o}\left(Z|Y,X_{1},...,X_{n},X_{n+1}\right) & =l\left(Z|Y\uplus X_{1}\uplus...\uplus X_{n}\uplus X_{n+1}\right).
\end{align*}

We have to prove two cases: $X_{n+1}=\emptyset$ and $X_{n+1}=\left\{ x\right\} $.
For $X_{n+1}=\emptyset$, we have that $Z_{n+1}=\emptyset$ so that
$t\left(Z_{n+1}|X_{n+1}\right)\neq0$. Therefore, 
\begin{align*}
 & l_{o}\left(Z|Y,X_{1},...,X_{n},\emptyset\right)\\
 & \quad=\sum_{Z_{1}\uplus...\uplus Z_{n}\uplus Z^{y}=Z}l\left(Z^{y}|Y\right)\prod_{i=1}^{n}t\left(Z_{i}|X_{i}\right)\\
 & \quad=l_{o}\left(Z|Y,X_{1},...,X_{n}\right)\\
 & \quad=l\left(Z|X\uplus\emptyset\right)
\end{align*}
where $X=Y\uplus X_{1}\uplus...\uplus X_{n}$. This proves the first
case. 

For $X_{n+1}=\left\{ x\right\} $, we have
\begin{align*}
 & l_{o}\left(Z|Y,X_{1},...,X_{n},\left\{ x\right\} \right)\\
 & =\sum_{Z_{1}...\uplus Z_{n}\uplus Z_{n+1}\uplus Z^{y}=Z}l\left(Z^{y}|Y\right)t\left(Z_{i}|\left\{ x\right\} \right)\prod_{i=1}^{n}t\left(Z_{i}|X_{i}\right)\\
 & =t\left(\emptyset|\left\{ x\right\} \right)\sum_{Z_{1}...\uplus Z_{n}\uplus Z^{y}=Z}l\left(Z^{y}|Y\right)t\left(Z_{i}|\left\{ x\right\} \right)\prod_{i=1}^{n}t\left(Z_{i}|X_{i}\right)\\
 & +\sum_{z\in Z}t\left(\left\{ z\right\} |\left\{ x\right\} \right)\sum_{Z_{1}\uplus...\uplus Z_{n}\uplus Z^{y}=Z\setminus\left\{ z\right\} }l\left(Z^{y}|Y\right)t\left(Z_{i}|\left\{ x\right\} \right)\\
 & \quad\times\prod_{i=1}^{n}t\left(Z_{i}|X_{i}\right)\\
 & =\left(1-p_{d}\left(x\right)\right)l\left(Z|X\right)\\
 & +p_{d}\left(x\right)\sum_{z\in Z}l\left(z|x\right)l\left(Z\setminus\left\{ z\right\} |X\right)\\
 & =l\left(Z|X\uplus\left\{ x\right\} \right)
\end{align*}
where $X=Y\uplus X_{1}\uplus...\uplus X_{n}$. This proves the second
case.

\section{\label{sec:AppendixC_2_multiBernoulli_mixture_birth}}

In this appendix, we prove the prediction step of Corollary \ref{cor:MBM_conjugacy}.
We consider that the new born targets follow an MBM with parameters
\begin{align}
f_{b}^{mbm}\left(Y\right) & \propto\sum_{j_{b}}\sum_{Y_{1}\uplus...\uplus Y_{n_{b}}=Y}\prod_{i_{b}=1}^{n_{b}}w_{j_{b},i_{b}}^{b}f_{j_{b},i_{b}}^{b}\left(Y_{i}\right).
\end{align}
As indicated in Section \ref{subsec:Prediction-conjugate-prior},
the predicted density of the survival targets when the Poisson intensity
is zero is an MBM. We denote the parameters of this MBM as in (\ref{eq:multiBernoullimixture_multiplicative}).
Then, the output of the prediction step is the multi-target density
of the union of the survival targets and the new born targets, which
can be computed using the convolution formula \cite[Eq. (4.17)]{Mahler_book14}
\begin{align*}
f_{pred}\left(W\right) & =\sum_{X\uplus Y=W}f^{mbm}\left(X\right)f_{b}^{mbm}\left(Y\right)\\
 & \propto\sum_{X\uplus Y=W}\left[\sum_{j}\sum_{X_{1}\uplus...\uplus X_{n}=X}\prod_{i=1}^{n}w_{j,i}f_{j,i}\left(X_{i}\right)\right]\\
 & \quad\times\left[\sum_{j_{b}}\sum_{Y_{1}\uplus...\uplus Y_{n_{b}}=Y}\prod_{i_{b}=1}^{n_{b}}w_{j_{b},i_{b}}^{b}f_{j_{b},i_{b}}^{b}\left(Y_{i}\right)\right]\\
 & =\sum_{j}\sum_{j_{b}}\sum_{X_{1}\uplus...\uplus X_{n}\uplus Y_{1}\uplus...\uplus Y_{n_{b}}=W}\\
 & \quad\left[\prod_{i=1}^{n}w_{j,i}f_{j,i}\left(X_{i}\right)\right]\left[\prod_{i_{b}=1}^{n_{b}}w_{j_{b},i_{b}}^{b}f_{j_{b},i_{b}}^{b}\left(Y_{i}\right)\right]
\end{align*}
which corresponds to an MBM.

\section{\label{sec:AppendixD_deltaGLMB}}

In this appendix, we prove Proposition \ref{prop:GLMB_density_equivalence}.
We first prove how a labelled MBM, which contains the labelled MBM$_{01}$
as a particular case, can be written as a $\delta$-GLMB density.
We write (\ref{eq:labelled_MB_mixture}) as 
\begin{align}
f\left(X\right) & =\sum_{j}w_{j}\sum_{X_{1}\uplus...\uplus X_{n}=X}\prod_{i=1}^{n}f_{j,i}^{lb}\left(X_{i}\right)\label{eq:labelled_MBM_appendix}
\end{align}
where we have normalised the weights of the global hypotheses such
that $\sum_{j}w_{j}=1$ and $w_{j}\propto\prod_{i=1}^{n}w_{j,i}$.
Let $\mathbb{L}=\left\{ \ell_{1},...,\ell_{n}\right\} $ denote the
set with all the possible target labels according to the density (\ref{eq:labelled_MB_mixture}). 

Both the $\delta$-GLMB density and the labelled multi-Bernoulli mixture
are zero if 1) they are evaluated on a set that includes more than
one target with the same label, or 2) if they are evaluated on a set
that includes a target whose label does not belong to the label space
$\mathbb{L}$. Therefore, the case of interest is when we evaluate
the density with a set of targets with distinct labels that belong
to $\mathbb{L}$. We evaluate the labelled multi-Bernoulli mixture
(\ref{eq:labelled_MBM_appendix})  on a labelled set $\left\{ \left(x_{1},\ell_{a_{1}}\right),...,\left(x_{p},\ell_{a_{p}}\right)\right\} $
where $\ell_{a_{1}},...,\ell_{a_{p}}$ are $p$ distinct labels that
belong to $\mathbb{L}$. We also denote by $\ell_{a_{p+1}},...,\ell_{a_{n}}$
the rest of distinct labels in $\mathbb{L}$. As labels $\ell_{a_{1}},...,\ell_{a_{p}}$
are distinct, there is only one combination in the sum over $X_{1}\uplus...\uplus X_{n}=X$
that is non-zero. This yields
\begin{align}
 & f\left(\left\{ \left(x_{1},\ell_{a_{1}}\right),...,\left(x_{p},\ell_{a_{p}}\right)\right\} \right)\nonumber \\
 & =\sum_{j}w_{j}\left[\prod_{m=1}^{p}r_{j,a_{m}}p_{j,a_{m}}\left(x_{m}\right)\right]\prod_{i=p+1}^{n}\left(1-r_{j,a_{i}}\right).\label{eq:evaluation_lMBM}
\end{align}

We proceed to write this density in the $\delta$-GLMB form \cite{Vo14}.
We denote 
\begin{align}
w_{j}\left(\left\{ \ell_{a_{1}},...,\ell_{a_{p}}\right\} \right) & =w_{j}\left[\prod_{m=1}^{p}r_{j,a_{m}}\right]\prod_{i=p+1}^{n}\left(1-r_{j,a_{i}}\right).\label{eq:delta_GLMB_weight_prev}
\end{align}
In the $\delta$-GLMB filter, this weight is written as (see sentence
that contains Eq. (9) in \cite{Vo14})
\begin{align}
w_{j}\left(\left\{ \ell_{a_{1}},...,\ell_{a_{p}}\right\} \right) & =\sum_{I\subseteq\mathbb{L}}w_{j}\left(I\right)\delta_{I}\left(\left\{ \ell_{a_{1}},...,\ell_{a_{p}}\right\} \right),\label{eq:delta_GLMB_weight}
\end{align}
where \cite{Vo14}
\begin{align*}
\delta_{I}\left(L\right) & \triangleq\begin{cases}
1 & \mathrm{if}\;I=L\\
0 & \mathrm{otherwise}
\end{cases}
\end{align*}
and it can be verified that $\sum_{j}\sum_{I\subseteq\mathbb{L}}w_{j}\left(I\right)=1$.
The previous step is direct, as there is only one summand in (\ref{eq:delta_GLMB_weight})
that is different from zero, which corresponds to (\ref{eq:delta_GLMB_weight_prev}).
Following \cite{Vo14}, we also denote $p_{\xi}\left(x,\ell\right)=p_{\xi,i\left(\ell\right)}\left(x\right)$
where $i\left(\ell\right)=i$ such that $\ell=\ell_{i}$ and index
$j$ is denoted as $\xi$. Substituting this notation into (\ref{eq:evaluation_lMBM}),
we find 
\begin{align}
 & f\left(\left\{ \left(x_{1},\ell_{a_{1}}\right),...,\left(x_{p},\ell_{a_{p}}\right)\right\} \right)\nonumber \\
 & =\sum_{\xi}\sum_{I\subseteq\mathbb{L}}w_{\xi}\left(I\right)\delta_{I}\left(\left\{ \ell_{a_{1}},...,\ell_{a_{p}}\right\} \right)\prod_{m=1}^{p}p_{\xi}\left(x_{m},\ell_{a_{m}}\right),\label{eq:delta_GLMB_density}
\end{align}
which corresponds to the $\delta$-GLMB density \cite[Eq. (9)]{Vo14}
evaluated on a set of targets with different labels. 

In order to finish the proof of Proposition \ref{prop:GLMB_density_equivalence},
we write a $\delta$-GLMB density as a labelled MBM with MBM$_{01}$
parameterisation. 

We consider that the label space is $\mathbb{L}=\left\{ \ell_{1},...,\ell_{n}\right\} $,
the $\delta$-GLMB single target densities are $p_{\xi}\left(\cdot,\ell_{i}\right)$
for all $\xi$ and $\ell_{i}\in\mathbb{L},$ and the global hypothesis
weights are $w_{\xi}\left(I\right)$ for $I\subseteq\mathbb{L}$.
In order to prove the equivalence, we evaluate a $\delta$-GLMB density
$f\left(\cdot\right)$ at $\left\{ \left(x_{1},\ell_{a_{1}}\right),...,\left(x_{p},\ell_{a_{p}}\right)\right\} $,
which is given by (\ref{eq:delta_GLMB_density}), with $\left\{ \ell_{a_{1}},...,\ell_{a_{p}}\right\} \subseteq\mathbb{L}$.
We also denote by $\ell_{a_{p+1}},...,\ell_{a_{n}}$ the rest of distinct
labels in $\mathbb{L}$. It should be noted that the pair $\left(\xi,I\right)$
represents a $\delta$-GLMB global hypothesis \cite{Vo14} and that,
in this global hypothesis, all targets whose label belongs to $I$
exist and the rest do not exist, which is represented by $\delta_{I}\left(\left\{ \ell_{a_{1}},...,\ell_{a_{p}}\right\} \right)$
in (\ref{eq:delta_GLMB_density}). For global hypothesis $\left(\xi,I\right)$,
this factor can be written as a product of existence probabilities,
which are either 0 or 1, as 
\begin{align}
 & \delta_{I}\left(\left\{ \ell_{a_{1}},...,\ell_{a_{p}}\right\} \right)\nonumber \\
 & \:=\left[\prod_{m=1}^{p}r_{\left(\xi,I\right),a_{m}}\right]\left[\prod_{i=p+1}^{n}\left(1-r_{\left(\xi,I\right),a_{i}}\right)\right],\label{eq:equivalence_target_existence}
\end{align}
where $r_{\left(\xi,I\right),a_{m}}=1$ if $\ell_{a_{m}}\in I$ and
$r_{\left(\xi,I\right),a_{m}}=0$ if $\ell_{a_{m}}\notin I$ for $m\in\left\{ 1,...,n\right\} $.
We can  write the two sums in (\ref{eq:delta_GLMB_density}) as one
sum over $j=\left(\xi,I\right)$ such that 
\begin{align}
 & f\left(\left\{ \left(x_{1},\ell_{a_{1}}\right),...,\left(x_{p},\ell_{a_{p}}\right)\right\} \right)\nonumber \\
 & =\sum_{j}w_{j}\left[\prod_{m=1}^{p}r_{j,a_{m}}\right]\left[\prod_{i=p+1}^{n}\left(1-r_{j,a_{i}}\right)\right]\prod_{m=1}^{p}p_{j,a_{m}}\left(x_{m}\right)\label{eq:delta_GLMB_density2}
\end{align}
where $p_{\left(\xi,I\right),a_{m}}\left(\cdot\right)=p_{\xi}\left(\cdot,\ell_{a_{m}}\right)$,
$w_{\left(\xi,I\right)}=w_{\xi}\left(I\right)$. It should be noted
that, in the $\delta$-GLMB density, we have $\sum_{\xi}\sum_{I\subseteq\mathbb{L}}w_{\xi}\left(I\right)=1$,
which implies that $\sum_{j}w_{j}=1$, as required. Also note that
$r_{j,i}$ is the existence probability of Bernoulli component $i$,
with label $\ell_{i}$, and global hypothesis $j$, which is either
0 or 1. Equation (\ref{eq:delta_GLMB_density2}) corresponds to the
evaluation of a labelled multi-Bernoulli mixture, see Equation (\ref{eq:evaluation_lMBM}).
In particular, the resulting global hypotheses (mixture components)
of the $\delta$-GLMB density are equivalent to the global hypotheses
in an MBM$_{01}$ parameterisation, which have deterministic target
existence. 

\bibliographystyle{IEEEtran}
\bibliography{9E__Trabajo_Angel_Mis_articulos_PMBM_Targets_Accepted_Referencias}

\end{document}